\documentclass[pdflatex, sn-mathphys-num]{sn-jnl}

\usepackage{graphicx}%
\usepackage{multirow}%
\usepackage{amsmath,amssymb,amsfonts}%
\usepackage{amsthm}%
\usepackage{mathrsfs}%
\usepackage[title]{appendix}%
\usepackage{xcolor}%
\usepackage{textcomp}%
\usepackage{manyfoot}%
\usepackage{booktabs}%
\usepackage{algorithm}%
\usepackage{algorithmicx}%
\usepackage{algpseudocode}%
\usepackage{listings}%
\usepackage[]{todonotes}
\usepackage{subcaption}
\usepackage[nolist]{acronym}
\usepackage{verbatim}

\begin{acronym}[ECU]
\acro{dsc}[DSC]{Dice Similarty Coefficient}
\acro{asd}[ASD]{Average Surface Distance}
\acro{tre}[TRE]{Target Registration Error}

\acro{cnn}[CNN]{Convoutional Neural Network}
\acro{gnn}[GNN]{Graph Neural Network}
\acro{mlp}[MLP]{Multi-Layer Perceptron}
\acro{tpe}[TPE]{Tree Structured Parzen Estimator}
\acro{jsrt}[JSRT]{Japanese Society of Radiological Technology}
\acro{graz}[GRAZPEDWRI-DX]{Graz Pediatric Wrist - Digital X-ray}
\acro{heatreg}[HeatReg]{Heatmap Regression}
\end{acronym}

\begin{document}

\title[DenseSeg]{DenseSeg: Joint Learning for Semantic Segmentation and Landmark Detection Using Dense Image-to-Shape Representation}


\author*[1]{\fnm{Ron} \sur{Keuth}}\email{r.keuth@uni-luebeck.de}

\author[2]{\fnm{Lasse} \sur{Hansen}}

\author[3]{\fnm{Maren} \sur{Balks}}

\author[3]{\fnm{Ronja} \sur{Jäger}}

\author[3]{\fnm{Anne-Nele} \sur{Schröder}}

\author[3]{\fnm{Ludger} \sur{Tüshaus}}

\author[1]{\fnm{Mattias} \sur{Heinrich}}

\affil*[1]{\orgdiv{Medical Informatics}, \orgname{University of L\"ubeck}, \orgaddress{\street{Ratzeburger Allee 160}, \postcode{23562} \city{L\"ubeck}, \country{Germany}}}

\affil[2]{\orgname{EchoScout GmbH}, \orgaddress{\street{Maria-Goeppert-Str. 3}, \postcode{23562} \city{L\"ubeck}, \country{Germany}}}

\affil[3]{\orgdiv{Paediatric Surgery}, \orgname{University Hospital Schleswig-Holstein}, \orgaddress{\street{Ratzeburger Allee 160}, \postcode{23562} \city{L\"ubeck}, \country{Germany}}}


\abstract{\textbf{Purpose:} Semantic segmentation and landmark detection are fundamental tasks of medical image processing, facilitating further analysis of anatomical objects.
Although deep learning-based pixel-wise classification has set a new-state-of-the-art for segmentation, it falls short in landmark detection, a strength of shape-based approaches.

\textbf{Methods:} In this work, we propose a dense image-to-shape representation that enables the joint learning of landmarks and semantic segmentation by employing a fully convolutional architecture.
Our method intuitively allows the extraction of arbitrary landmarks due to its representation of anatomical correspondences. We benchmark our method against the state-of-the-art for semantic segmentation (nnUNet), a shape-based approach employing geometric deep learning and a convolutional neural network-based method for landmark detection.

\textbf{Results:} We evaluate our method on two medical dataset: one common benchmark featuring the lungs, heart, and clavicle from thorax X-rays, and another with 17 different bones in the paediatric wrist. While our method is on pair with the landmark detection baseline in the thorax setting (error in mm of $2.6\pm0.9$ vs $2.7\pm0.9$), it substantially surpassed it in the more complex wrist setting ($1.1\pm0.6$ vs $1.9\pm0.5$).

\textbf{Conclusion:} We demonstrate that dense geometric shape representation is beneficial for challenging landmark detection tasks and outperforms previous state-of-the-art using heatmap regression. While it does not require explicit training on the landmarks themselves, allowing for the addition of new landmarks without necessitating retraining.}




\maketitle






\section{Introduction and Related Work}
Semantic segmentation and landmark detection are fundamental tasks of medical image processing, introducing representations for further analysis of the image object.
While semantic segmentation identifies an image object, landmarks represent anatomical points, providing correspondences between images.
In the contemporary era, both of these tasks can be effectively tackled using deep learning methodologies.

\subsection{Landmark Detection}
For landmark detection, the variety of supervised learning approaches utilizing \acp{cnn}, can be clustered into three categories:
1) \textbf{Direct regression}, where the model directly regresses the landmarks' coordinates \cite{Ranjan_HyperFace} or utilize a cascade system refining previous predictions iteratively \cite{Sun_FaceCascade}.
However, a potential drawback of this approach is that it requires the extraction of global numbers from the local \ac{cnn}'s feature map, which may not fully leverage the local characteristics of \acp{cnn}.
While 2) \textbf{pixelwise classification} overcomes this by estimating the likelihood of each pixel being the landmark position \cite{He2017MaskR}, it can potentially become an ill-posed problem due to its loss formulation (e.g., cross entropy) introducing a "one-against-all" principle.
3) \textbf{Heatmap regression} \cite{leibe_heatmap_regression_2016} is closely related to the previous approach, but addresses its drawback by formulating the landmark detection as a heatmap where its values are proportional to the likelihood of each pixel being a landmark. This allows for a more flexible representation, where multiple neighbouring pixels can have high likelihoods without incurring a penalty.

\subsection{Semantic Segmentation}
The UNet \cite{Ronneberger2015UNet} has introduced a new state-of-the for semantic segmentation and offers with its self-adapting nnU-Net framework \cite{isensee_nnu-net_2021} superb out-of-the-box segmentation on a variety of biological and medical tasks.
However, the UNet's utilization of pixel-wise classification lacks a representation of landmarks, which is natively present in a shape-based approach, where a shape model is aligned from a known to a new image comprising anatomical landmarks.
While in the past, shape-based approaches has been dominated by active shape models, nowadays, the main focus lies on deep learning-based approaches to overcome the expensive iteratively fitting of shape models.
With geometric deep learning, the explicit learning of geometric features has been introduced, and newer work combines image-based \ac{cnn} backbones with \acp{gnn} to tackle the problem of segmentation:
In PolyTransform \cite{liang_polytransform_2020}, the contour of a predicted segmentation mask is refined by a transformer block, and \cite{gaggion_hybridnet_2023} combines a \ac{cnn} encoder and a graph convolutional decoder with a new type of image-to-graph skip connections.
\cite{keuth_combining_2024} combines the PointTransformer with a \ac{cnn} to estimate the displacements of an initial shape, warping it to the image object in a one-shot manner.
In \cite{Neural_Body_Fitting} an integration of a differentiable shape model into a \ac{cnn} has been proposed providing an end-to-end training.
However, most approaches have not reached the performance of \ac{cnn}-based pixelwise classification.\par

\subsection{Dense Representation for Landmark Detection}
Motivated by this, we propose a method allowing to utilize state-of-the-art \ac{cnn} architectures for semantic segmentation by also providing the ability of landmark detection.
With this multitasking, our model takes advantage of the close relationship of these tasks, which has been already proven in \cite{Jackson2016ACC, Takikawa2019GatedSCNNGS, Gler2018DensePoseDH, He2017MaskR}.
We achieve this by introducing a dense image-to-shape representation, which enables the extraction of arbitrary landmarks on unseen images.
Different to conventional shape models, our approach does not require an initial shape.
With our representation being a normalised Cartesian coordinate system called $uv$ coordinates, our work is closely related and inspired by DenseReg \cite{Gler2016DenseRegFC}.
DenseReg employs this $uv$-maps on human faces, demonstrating its potential as a foundational representation for subsequent computer vision tasks like segmentation.
With DensePose \cite{Gler2018DensePoseDH} the dense correspondences of $uv$ coordinates were introduced for the whole human body by jointly learning the task of instance segmentation of human as well as the semantic segmentation of their human parts.
Later, the use of such a normalized presentation found its application in scene representation \cite{Wang2019NormalizedOC} and generation of accurate 3D models from a single image view, where an important intermediate step includes a representation in $uv$ coordinates.
The polar coordinate system offers an alternative representation method, where the center of the object serves as the origin and multiple rays, each with a regular angular offset, extend towards the object's boundaries.
While this representation has been proven to be effective in the context of instance segmentation of common objects \cite{hurtik_poly-yolo_2022, Xie2019PolarMaskSS} and also in the biomedical domain \cite{Schmidt2018CellStar}, it has not been utilized for correspondence representation yet.
Besides the explicit formulation, \cite{Thewlis2017UnsupervisedOL} proposed an unsupervised approach, where a \ac{cnn} finds a canonical latent space representing semantic-consistent coordinates and thus provides landmarks on an unseen image of the same object type.
However, this approach necessitates a large dataset of the same object, limiting its use in the medical field where data, especially for rare diseases, is often scarce.

\section{Methods}
\subsection{Problem Definition}\label{sec:uv_definition}
In semantic segmentation, we are given an image $\mathbf{I}\in\mathbb{R}^{H\times W}$ and aim to learn a function $f_s(\mathbf{I})$ that estimates the segmentation mask $\hat{\mathbf{S}}\in[0, 1]^{|\Omega|\times H \times W}$, which holds all likelihoods $\hat{p}(\mathbf{i}|\omega)$ of each pixel $\mathbf{i}\in\mathbf{I}$ to belong to the class $\omega\in\Omega$.\par
In our setup, we also estimate a bijective function $\varphi:\mathbb{R}^{H\times W\times2}\to\mathbb{R}^{H\times W\times2}$ that warps a template $\mathbf{T}\in\mathbb{R}^{H\times W}$ to a given image $\mathbf{I}$ by introducing a deformation field such that:
\begin{equation}
    \mathbf{I}(x, y)\approx\mathbf{T}(\varphi(h,w)).
\end{equation}
However, similar to the DenseReg \cite{Gler2016DenseRegFC} approach, we reformulate $\varphi$ in a way that it does not describe a relative displacement field but an absolute one introducing a one-to-one mapping of each pixel in $\mathbf{I}$ to its corresponding anatomical position in $\mathbf{T}$.
Given that the $\mathbf{T}$ is defined as being free of deformation, the anatomical position can be represented within a normalized 2D coordinate space, denoted as $\{u, v\}$, which falls within the range of $[-1, 1]$.
Consequently, we explicitly enforce $\varphi$ to describe anatomical correspondences, which is inherently beneficial for landmark detection.
We aim to solve both tasks by jointly learning $f_s$ and $f_\varphi\approx\varphi$.

\subsubsection{Generation of $uv$-Maps}

\begin{figure}
    \centering
    \includegraphics[width=.8\textwidth]{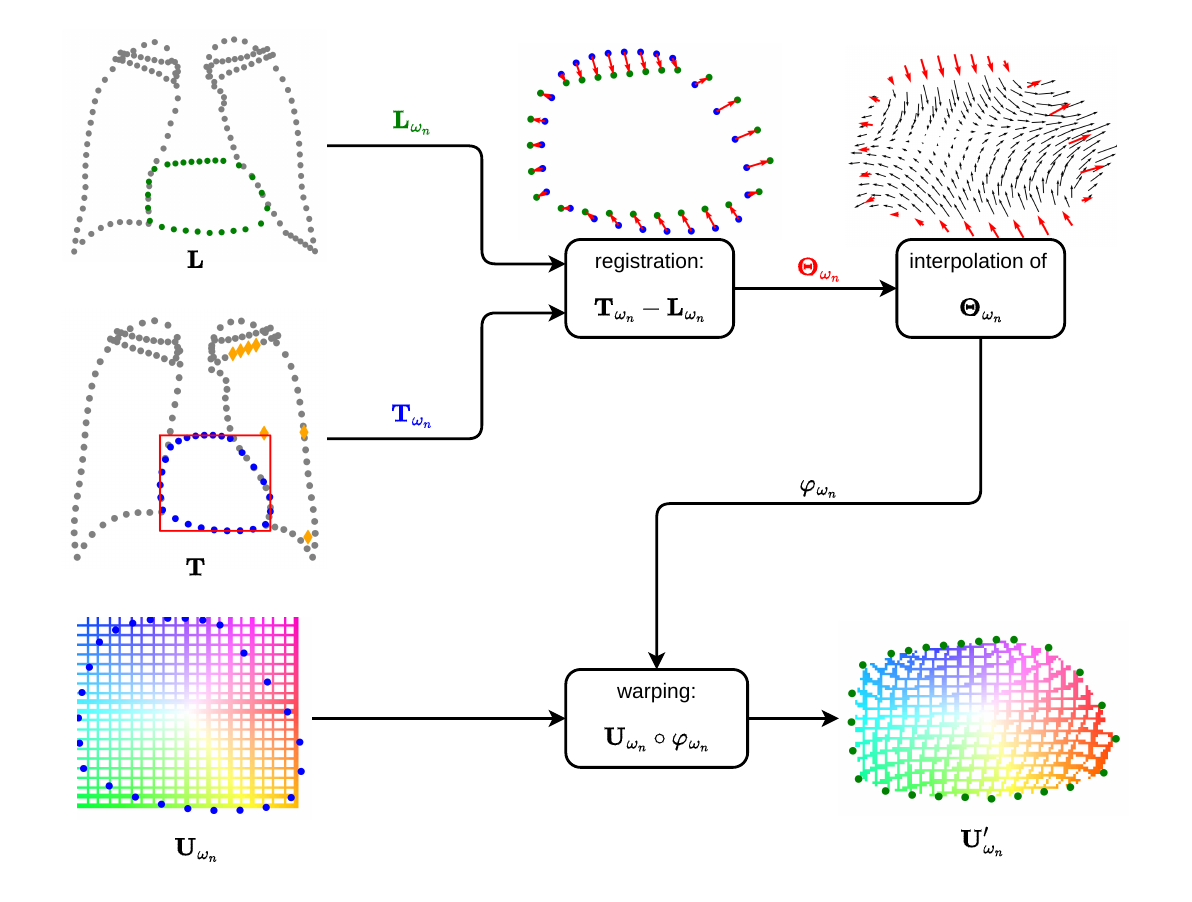}
    \caption{$uv$-Map generation: the corresponding landmarks $\mathbf{L}_{\omega_n}$ of an anatomical structure $\omega_n$ are aligned to a template $\mathbf{T}_{\omega_n}$ yielding a sparse displacement field $\mathbf{\Theta}_{\omega_n}$, which is interpolated to a dense one $\varphi_{\omega_n}$. The final step involves warping the template's $uv$-map $\mathbf{U}_{\omega_n}\circ\varphi_{\omega_n}$ to generate the $uv$-map $\mathbf{U}_{\omega_n}'$ that will be used for supervision during the training. $\textcolor{orange}{\blacklozenge}$ markers in $\mathbf{T}$ correspond to unknown landmarks not used in training (more details in Sec. \ref{sec:unknown_landmarks}).}
    \label{fig:uv_generation}
\end{figure}

We calculate the mean shape of all $N$ landmarks over the training split and use it as our deformation-free template $\mathbf{T}\in\mathbb{R}^{N\times2}$.
By indexing the pixels of the axis aligned bounding box of each anatomical structure $\omega_n$ (red box in Fig. \ref{fig:uv_generation}) in the $uv$ manner (see Sec. \ref{sec:uv_definition}), we obtain its identity $uv$-map $\mathbf{U}_{\omega_n}\in\mathbb{R}^{2\times H_n\times W_n}$, where $H_n$ and $W_n$ describing the region of the image covered by $\omega_n$.
Next, we align the landmarks $\mathbf{L}_{\omega_n}\in\mathbb{R}^{N\times2}$ to $\mathbf{T_{\omega_n}}$ using an affine transformation (no shearing) by utilizing the Umeyama algorithm \cite{Umeyama}, which estimates the optimal translation, rotation, and scaling factors to align two sets of points.
This is achieved through a least-squares approach based on the singular value decomposition of their cross-covariance matrix.
After obtaining the sparse displacement field $\mathbf{\Theta}_{\omega_n}=\mathbf{T}_{\omega_n}-\mathbf{L}_{\omega_n}$, we convert it to a dense field $\mathbf{\varphi}_{\omega_n}\in\mathbb{R}^{H_n\times W_n}$ employing bilinear interpolation, which is well-suited due to the smooth nature of $uv$ maps (see motivation for the use of total variation in Sec. \ref{seq:loss_fnc}). 
Finally, we obtain our $uv$-map $\mathbf{U}_{\omega_n}'$ by warping the identity map:
\begin{equation}
    \mathbf{U}_{\omega_n}'(h_n, w_n) = \mathbf{U}_{\omega_n}(\varphi(h_n, w_n)).
\end{equation}
Please note that as a result of applying bilinear interpolation to $\mathbf{\Theta}_{\omega_n}$, we only acquire valid values within the convex hull formed by the landmarks $\mathbf{L}_{\omega_n}$.
Fig. \ref{fig:uv_generation} provides a visualisation of the entire process for the heart.

\subsubsection{Landmark Extraction from $uv$-Maps}\label{sec:uv_landmark_extraction}
To gather the position of a given landmark $\mathbf{l}\in\mathbb{R}^2$ from a predicted $uv$-map $\hat{\mathbf{U}}$, we first sample its $uv$ values from its known position in the identity $uv$-map $\mathbf{U}$ and find the element and thus the position holding the closest $uv$ value in $\hat{\mathbf{U}}$.
To enable subpixel sampling, we consider the $K=5$ closest points $\mathcal{K}=\{\mathbf{k}_i\}_{i=1}^K$ in $\hat{\mathbf{U}}$ and perform a linear combination of their position, where the weights are obtained by a $\texttt{softmin}$ transformation of their $uv$ value distance to the one of the given landmark:
\begin{equation}
    \hat{\mathbf{l}} = \sum_{\mathbf{k}_i\in\mathcal{K}}\frac{\exp(-||\hat{\mathbf{U}}(\mathbf{k}_i)-\mathbf{U}(\mathbf{l})||_2)}{\sum_{\mathbf{k}_j\in\mathcal{K}}\exp(-||\hat{\mathbf{U}}(\mathbf{k}_j)-\mathbf{U}(\mathbf{l})||_2)}\cdot\mathbf{k}_i.
\end{equation}

\subsubsection{Loss Function}\label{seq:loss_fnc}
We employ several losses for the supervised training of our model.
For the segmentation head $f_s$, we chose the binary cross entropy $\mathcal{L}_\text{BCE}$ and weighted it with the ratio of positive and negative pixels per class obtained over the training split to balance recall and precision.
The regression head $f_\varphi$ is optimized using the $L_1$-norm, which offers greater training stability compared to the $L_2$-norm in our setting, to minimize $\mathcal{L}_\varphi$ representing the difference between ground truth and predicted $uv$ values.
Due to the implementation of the bilinear interpolation of the sparse displacement field, we obtain only values in the convex hull of the landmarks (see Sec. \ref{sec:uv_definition}) and thus would lack supervision at its borders.
We overcome this by sampling the predicted $uv$ values at the landmarks' ground truth coordinates and compare them to the one of the template $T$, obtaining $\mathcal{L}_\text{LM}$, which is normalized by the number of landmarks associated with each class.
Since both losses, $\mathcal{L}_\varphi$ and $\mathcal{L}_\text{LM}$, operate on the $uv$ maps, we combine them by averaging and use a shared weighting factor, $\lambda_\varphi$ to control their contribution to the final loss.
Because a rapid value change in the $uv$-maps is unlikely, we introduce total variation as a regularisation $\mathcal{L}_\text{TV}$.
The final loss $\mathcal{L}$ is then a weighted sum $\sum_{x\in\{\text{BCE},\varphi,\text{TV}\}}\lambda_x=1$ of all introduced losses:
\begin{equation}\label{eq:loss_function}
    \mathcal{L} = \lambda_\text{BCE}\cdot\mathcal{L}_\text{BCE}+\frac{\lambda_\varphi}{2}\cdot(\mathcal{L}_\varphi + \mathcal{L}_\text{LM}) + \lambda_\text{TV}\cdot\mathcal{L}_\text{TV}.
\end{equation}

\section{Experiments}
\subsection{Datasets}
First, we train and test our methods on the \ac{jsrt} Dataset \cite{shiraishi_jsrt_2000}. The \ac{jsrt} consists of 247 chest X-rays with a resolution of $1024\times1024$ pixels and an isotropic pixel spacing of $0.175\,\text{mm}$. Each image comes with human expert landmark annotations ($N=166$) for the four anatomical structures of the right lung (44), left lung (50), heart (26), right clavicle (23), and left clavicle (23). For our experiments, we downsample the dataset to $256\times256$ pixels (matching our set-up in \cite{keuth_combining_2024}) and divide it into a custom split of 160 training and 87 test images.

Second, we establish a new landmark detection dataset of paediatric wrist bones based on the publicly available \acf{graz} \cite{nagy_pediatric_2022} dataset. Since the dataset only provides raw images of a large cohort with 20k wrist trauma radiographs of 6k children and adolescents and manually segmenting several small bones with corresponding landmarks would be extremely challenging, we propose a semi-automatic process. In our pre-processing, the approximately 10k lateral views are excluded (since it is impossible to annotate several overlapping bones in a side view) and we normalise the orientation to left laterality. A subset of 40 scans is manually annotated with pixel level segmentation labels by two experienced paediatric physicians (surgeon and radiologist), which serves as starting point for a semi-supervised label propagation with SAM-based mask refinements \cite{keuth2024samcarriesburden}. Next, we create corresponding landmarks by training an unbiased point cloud registration network that aligns 4850 surface points extracted from each segmentation contours using a two-step SegResNet with the DiVRoC loss \cite{heinrich2023chasing}. We select one scan as reference and uniformly subsample points on each of the 17 bones - in total 720. After manual quality control this yields 4830 radiographs with landmark annotations that are rescaled to $384\times224$ and split 60/40 into training and test sets.

\subsection{Bone Age Regression with Landmarks}
As an exemplary downstream task for automatic landmark detection in wrist bones, we explore the possibility to regress a child's bone age using a simple MLP-regressor.
Given the 2D coordinates of landmarks for all 17 bones, we specified an \ac{mlp} with three Linear-BatchNorm-ReLU blocks that work on each bone separately to extract features followed by two fully-connected layers that use a concatenation of those inputs for a classification of bone age into buckets of 4 months each.
The hidden dimension was chosen as 32 for the first part and 64 for the second with an input of up to 80 landmark coordinates per bone, which were centred around zero individually beforehand.
An ensemble of five of these \acp{mlp} were trained for 75 epochs using Adam and combined during inference to avoid overfitting.
During training both predicted and ground truth landmarks were included, during test only the predicted ones.

\subsection{Addition of New Landmarks without Retraining}\label{sec:unknown_landmarks}
To showcase the flexibility of the $uv$ paradigm in incorporating new landmarks without requiring retraining, we examine four different locations on the left lung within the \ac{jsrt} setting. These landmarks include four points along the lower contour of the right clavicle, the lung center, one on the outer lung contour, and one at the center of the lung apex (indicated by the $\textcolor{orange}{\blacklozenge}$ markers in $\mathbf{T}$ in Fig. \ref{fig:uv_generation}). For each of these new landmarks, we sample their corresponding $uv$ values from the left lung's template $\mathbf{T}$. During inference, we then determine their positions from the predicted $uv$ map, as outlined in Sec. \ref{sec:uv_landmark_extraction}.

\subsection{Network Architecture} \label{sec:architecture}
\begin{figure}
    \centering
    \includegraphics[width=\textwidth]{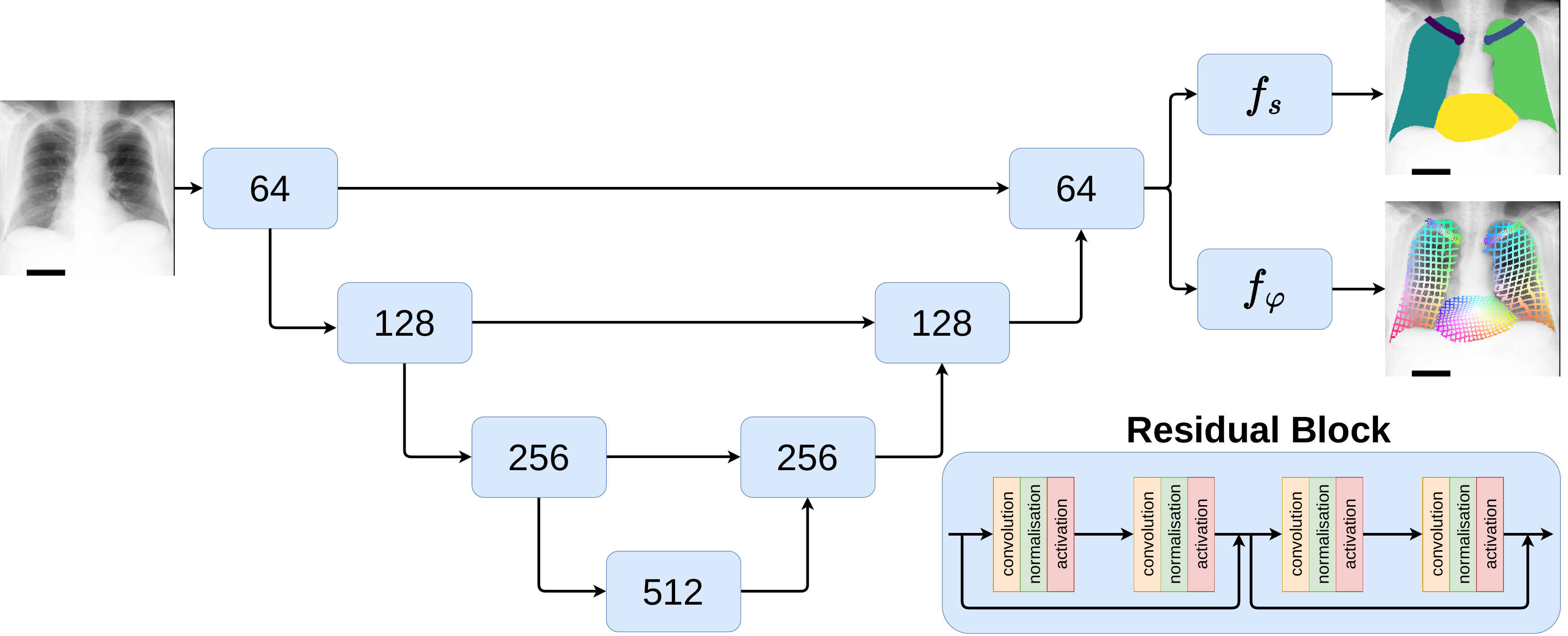}
    \caption{Two-head UNet for joint semantic segmentation $f_s$ and $uv$-mapping $f_\varphi$. A blue box symbolizes a residual block that contains two residual units with the label indicating the number of its output channels. The UNet extracts image features, thereby creating a canonical feature space both heads can utilize for segmentation and $uv$-mapping, respectively.}
    \label{fig:architecture}
\end{figure}
We utilize a UNet \cite{Ronneberger2015UNet} (MONAI implementation) with two heads to implement the segmentation $f_s$ and $uv$-mapping $f_\varphi$ (Fig. \ref{fig:architecture}).
The UNet itself consists of 6.4 million parameters, where the first convolution block extracts 64 channels, which are then doubled at each of the four encoder steps by halving its spatial dimensions. Each encoder as well as decoder step consists of two residual units \cite{He2016}.
With this, the UNet extracts 64 canonical features for each pixel, which are further processed by each head mapping it into an individual latent space of dimension 64. Each head consists of two residual units and 150 000 parameters in total.
Intuitively, we implement $f_\varphi$ by splitting it into a set of functions $\{f_\varphi^{\omega_n}\}_{n=1}^{|\Omega|}$, learning an individual function $f_\varphi^{\omega_n}$ for each anatomical structure $\omega_n$.
Because $f_\varphi^{\omega_n}$ is only defined in the area of its anatomical structure, we masked its output elsewhere, utilizing the ground truth segmentation mask during training and the predicted segmentation mask of $f_s$ for inference.

\subsection{Training and Hyperparameter Search}\label{sec:hyperparams}
The loss (see Eq. \ref{eq:loss_function}) is minimized by an ADAM optimizer with its proposed default parameters.
We reduce the initial learning rate by a factor of 100 with cosine annealing over 100 epochs.
During training, we us random affine transformation to synthetically increase the variability of our training data sampling ration from $\mathcal{U}(0^\circ, 10^\circ)$, relative translation for both image axis from $\mathcal{U}(0, 0.1)$ and relative scale factor from $\mathcal{U}(0.85, 1.15)$.
We do not use any test time augmentation.\par
The different $\lambda$ of Eq. \ref{eq:loss_function} are optimized by the \ac{tpe} approach (Optuna implementation) trying 150 different combinations also called trials.
We then pick from all trials describing the Pareto front, the one that yields a good compromise between the segmentation and regression  performance, resulting in $\lambda_\text{BCE}=0.13, \lambda_\varphi=0.66$ and $\lambda_\text{TV}=0.2$, where each $\lambda \in [0, 1]$.

\subsection{Comparison Methods}
For comparison, we chose our previous work \cite{keuth_combining_2024} combining image- and geometric deep learning for shape regression.
Here, we utilize an ImageNet-pretrained ResNet18 \cite{He2016} by cropping it to the first 12 layers and replacing the last two convolutions with dilated ones, obtaining a higher resolution feature map of $32 \times 32$ pixels with 64 channels.
We then employ the PointTransformer \cite{zhao_point_2021} to estimate the displacement of an initial shape (randomly picked from another training sample) by bilinear sampling features from the CNN feature map at the initial shape's coordinates.\par
As the state-of-the in semantic segmentation, we include the nnU-Net framework \cite{isensee_nnu-net_2021}. With it, we train two U-Nets for the \ac{jsrt} dataset without the default cross validation (fold \texttt{all} option) to avoid overlapping of lungs and clavicles.\par
As a comparison for landmark detection utilizing dense representation, we implement the \acf{heatreg} approach \cite{leibe_heatmap_regression_2016} using the same UNet architecture as in our method (see Sec. \ref{sec:architecture}).
Here each landmark' position $(h_l, w_l)$ is represented as an individual heatmap $\mathbf{H}_l\in\mathbb{R}^{H\times W}$ generated with a Gaussian kernel 
\begin{equation}
    \mathbf{H}_l(h, w)=\alpha\cdot\exp\left(\frac{((h_l-h)^2+(w_l-w)^2)}{2\sigma^2}\right).
\end{equation}
The model then regresses for each landmark its heatmap by minimizing the squared $L_2$-norm $\mathcal{L}_\text{H}$.
To compare the combination of landmark detection with semantic segmentation, we extend the \ac{heatreg} approach by incorporating the same segmentation head used in our method (see Sec. \ref{sec:architecture}), which we refer to as HeatRegSeg. During multitask training, we minimize the loss function $\mathcal{L} = \lambda \cdot \mathcal{L}_\text{BCE} + (1-\lambda) \cdot \mathcal{L}_\text{H}$, where $\mathcal{L}_\text{BCE}$ represents the loss for the segmentation task (see Sec. \ref{seq:loss_fnc}).
For \ac{heatreg} and HeatRegSeg, we employ the same hyperparameters and data augmentation as in Sec. \ref{sec:hyperparams} and optimize $\alpha=44\in[1,50]$, $\sigma=8\in[1, 10]$ and also for HeatRegSeg's $\lambda=0.38\in[0, 1]$ with the \ac{tpe}.

\section{Results and Discussion}
\begin{figure}
    \centering
    \begin{subfigure}{\textwidth}
        \centering
        \caption*{\textbf{Ours}}
        \begin{subfigure}{.3\textwidth}
            \includegraphics[width=\textwidth]{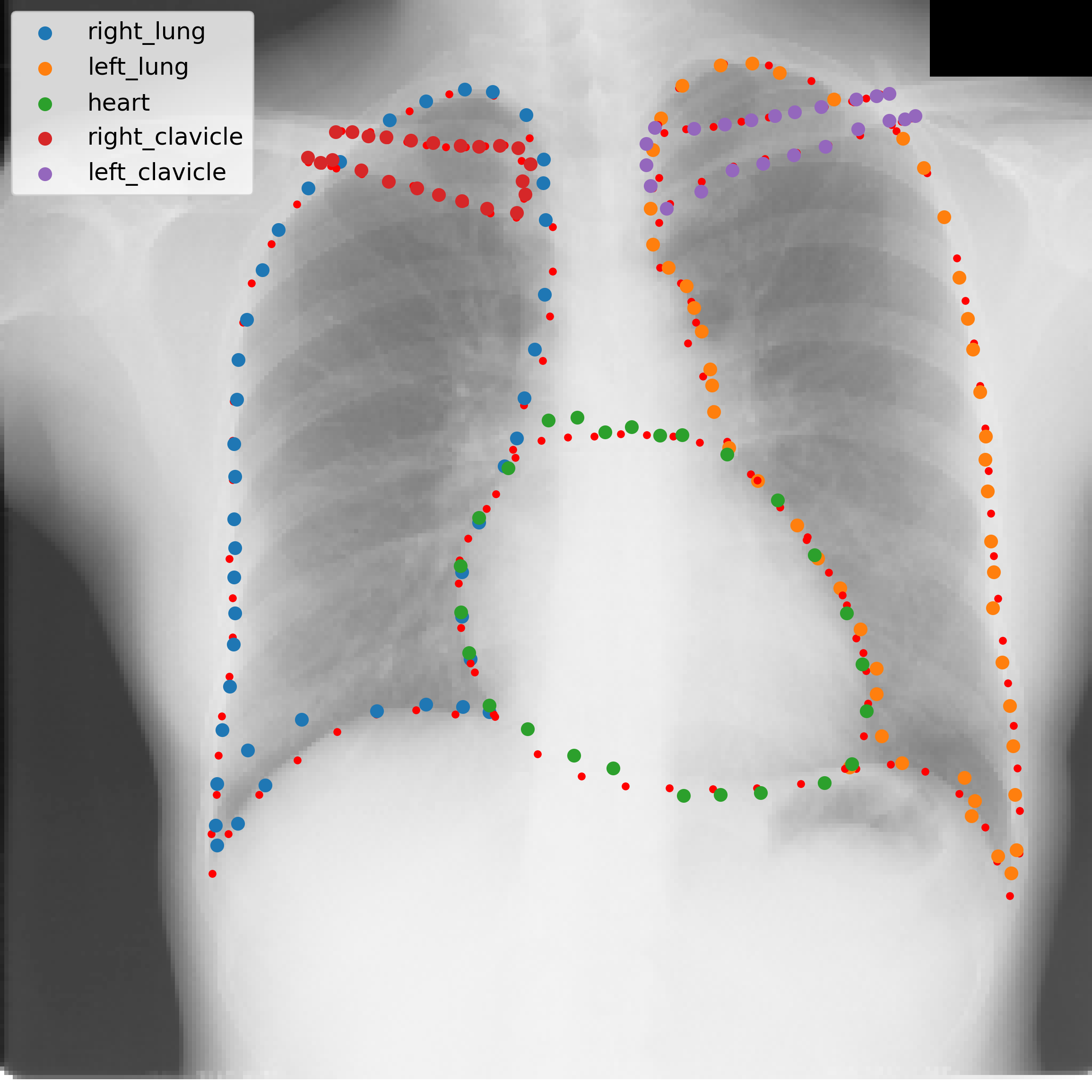}
            \caption{\begin{tabular}{rl}
                \acs{tre}: & 4.0 ± 2.0\\
                \acs{asd}: & 1.5 ± 0.5\\
            \end{tabular}}
        \end{subfigure}\hfill
        \begin{subfigure}{.3\textwidth}
            \includegraphics[width=\textwidth]{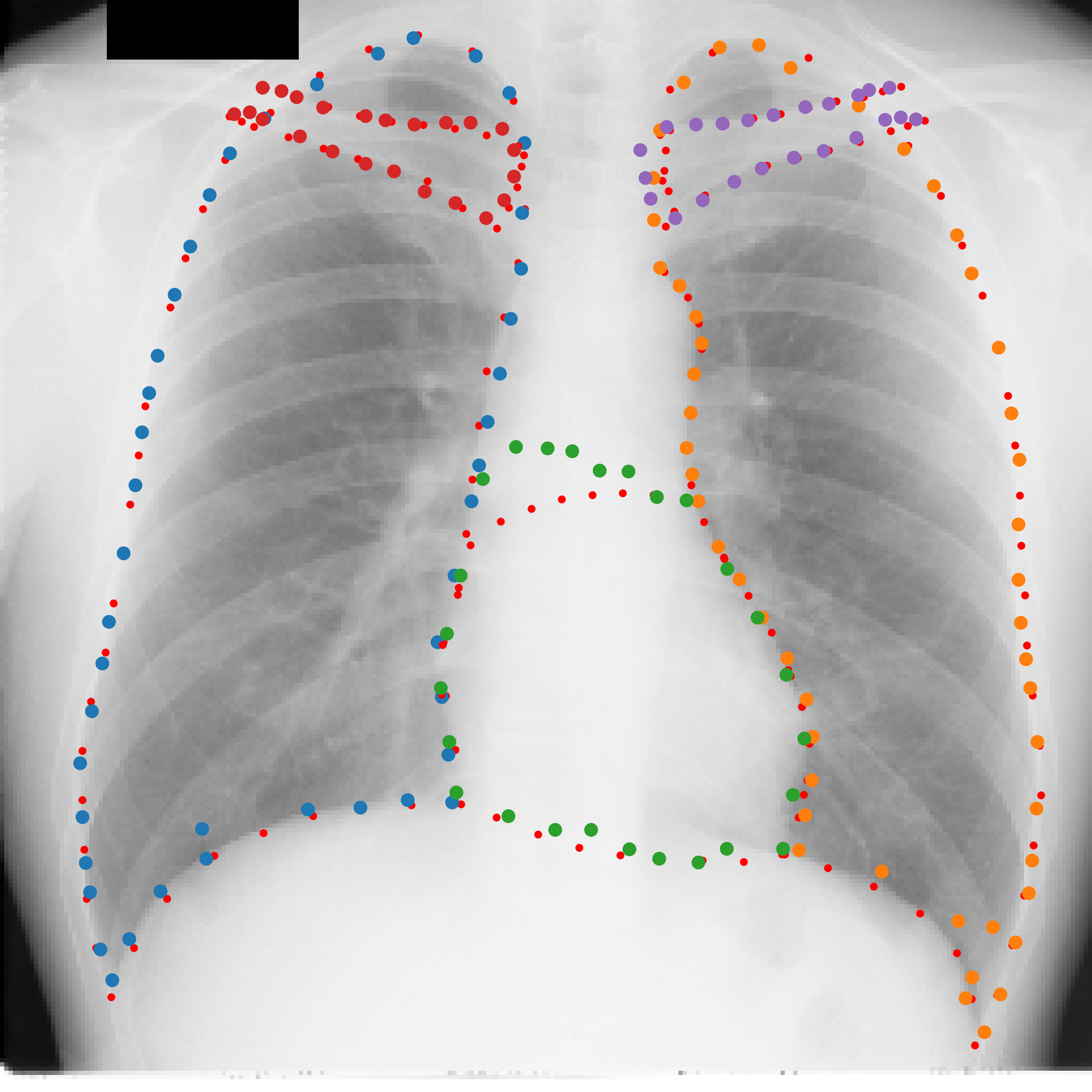}
            \caption{\begin{tabular}{rl}
                \acs{tre}: & 4.2 ± 1.4\\
                \acs{asd}: & 2.2 ± 1.1\\
            \end{tabular}}
            \label{fig:jsrt_heart_overestimation}
        \end{subfigure}\hfill
        \begin{subfigure}{.3\textwidth}
            \includegraphics[width=\textwidth]{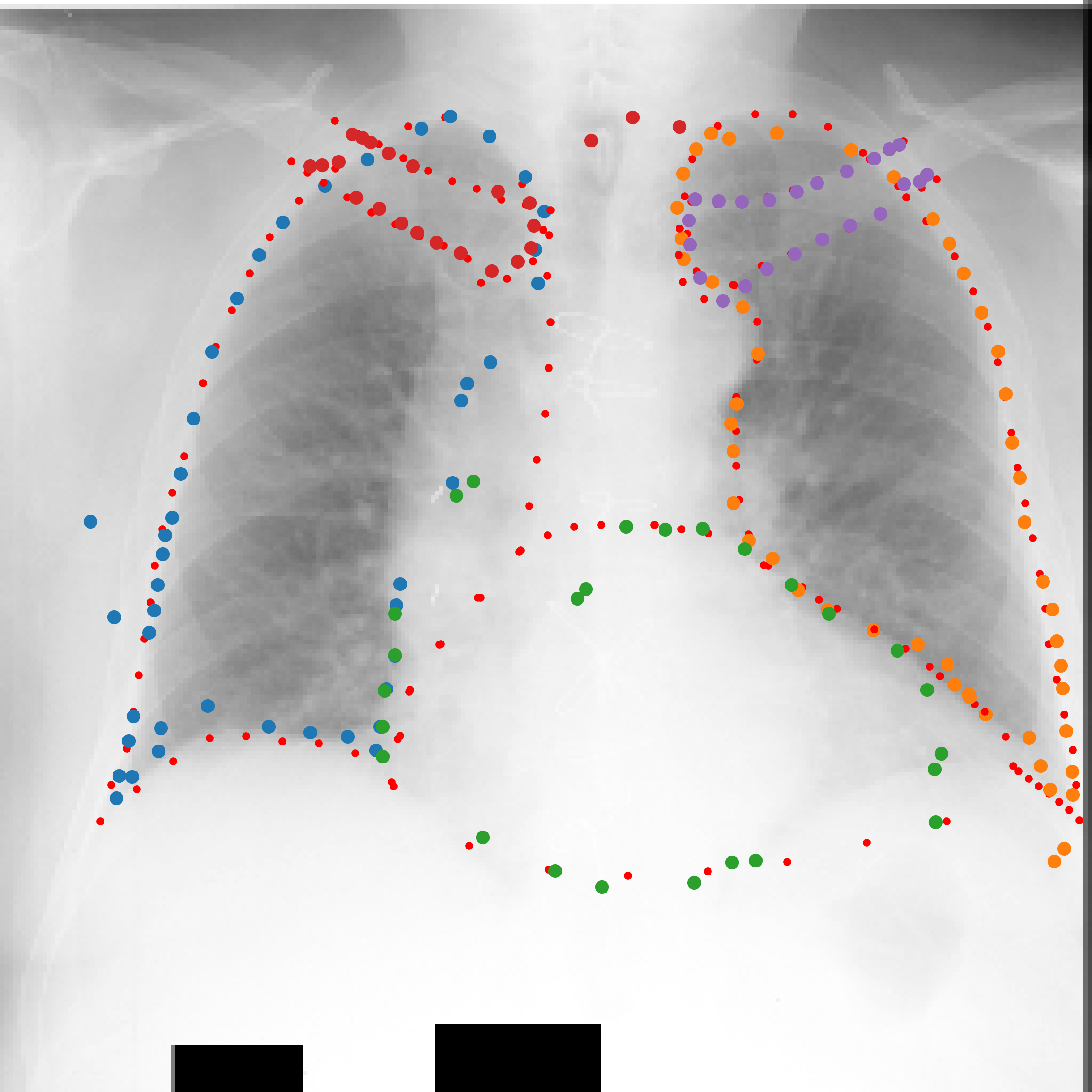}
            \caption{\begin{tabular}{rl}
                \acs{tre}: & 9.9 ± 4.9\\
                \acs{asd}: & 5.2 ± 2.7\\
            \end{tabular}}
        \end{subfigure}
    \end{subfigure}
    \begin{subfigure}{\textwidth}
        \centering
        \caption*{\textbf{Heatmap Regression}}
        \begin{subfigure}{.3\textwidth}
            \includegraphics[width=\textwidth]{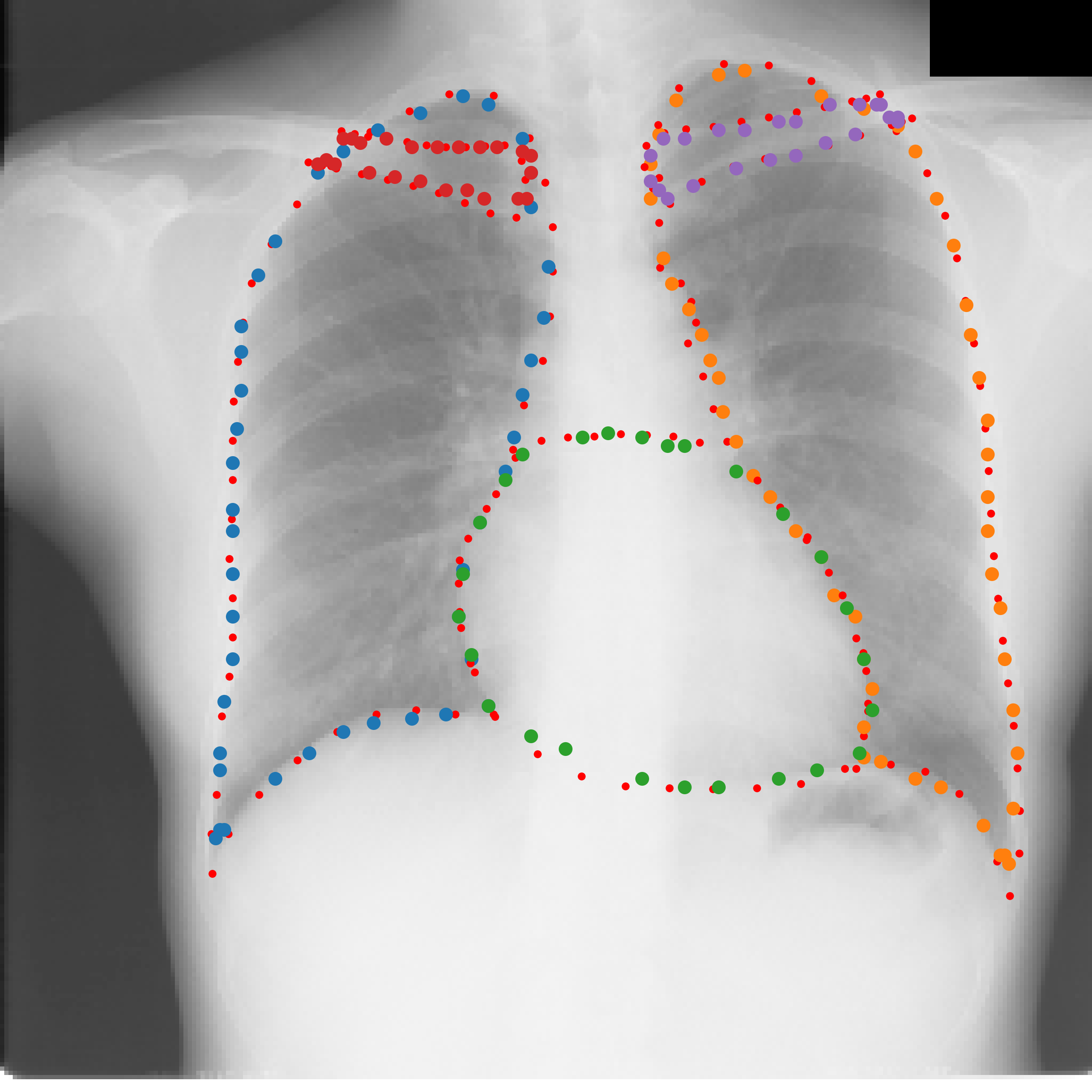}
            \caption{\begin{tabular}{rl}
                \acs{tre}: & 4.3 ± 1.3\\
                \acs{asd}: & 1.9 ± 0.2\\
            \end{tabular}}
        \end{subfigure}\hfill
        \begin{subfigure}{.3\textwidth}
            \includegraphics[width=\textwidth]{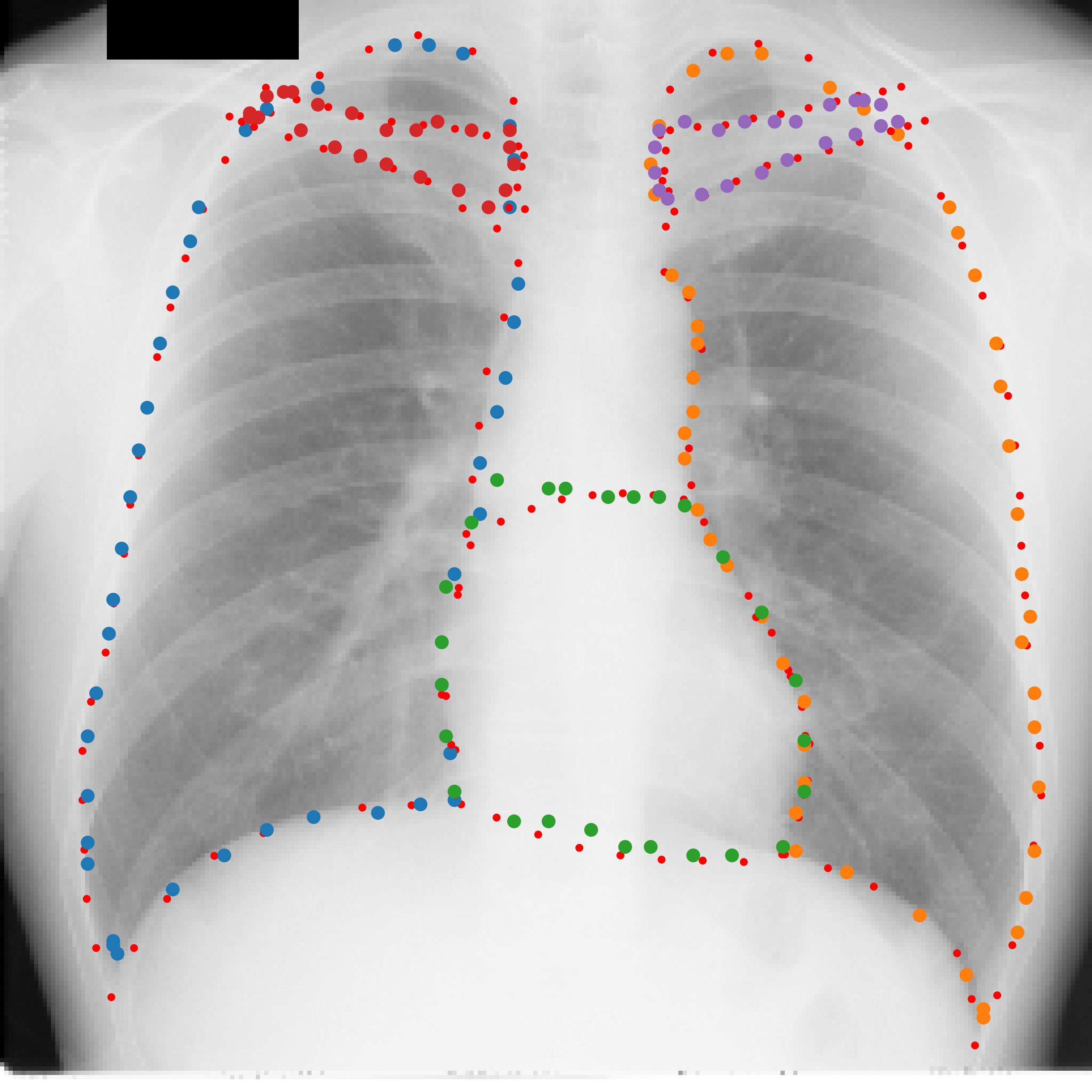}
            \caption{\begin{tabular}{rl}
                \acs{tre}: & 4.3 ± 0.4\\
                \acs{asd}: & 2.0 ± 0.4\\
            \end{tabular}}
        \end{subfigure}\hfill
        \begin{subfigure}{.3\textwidth}
            \includegraphics[width=\textwidth]{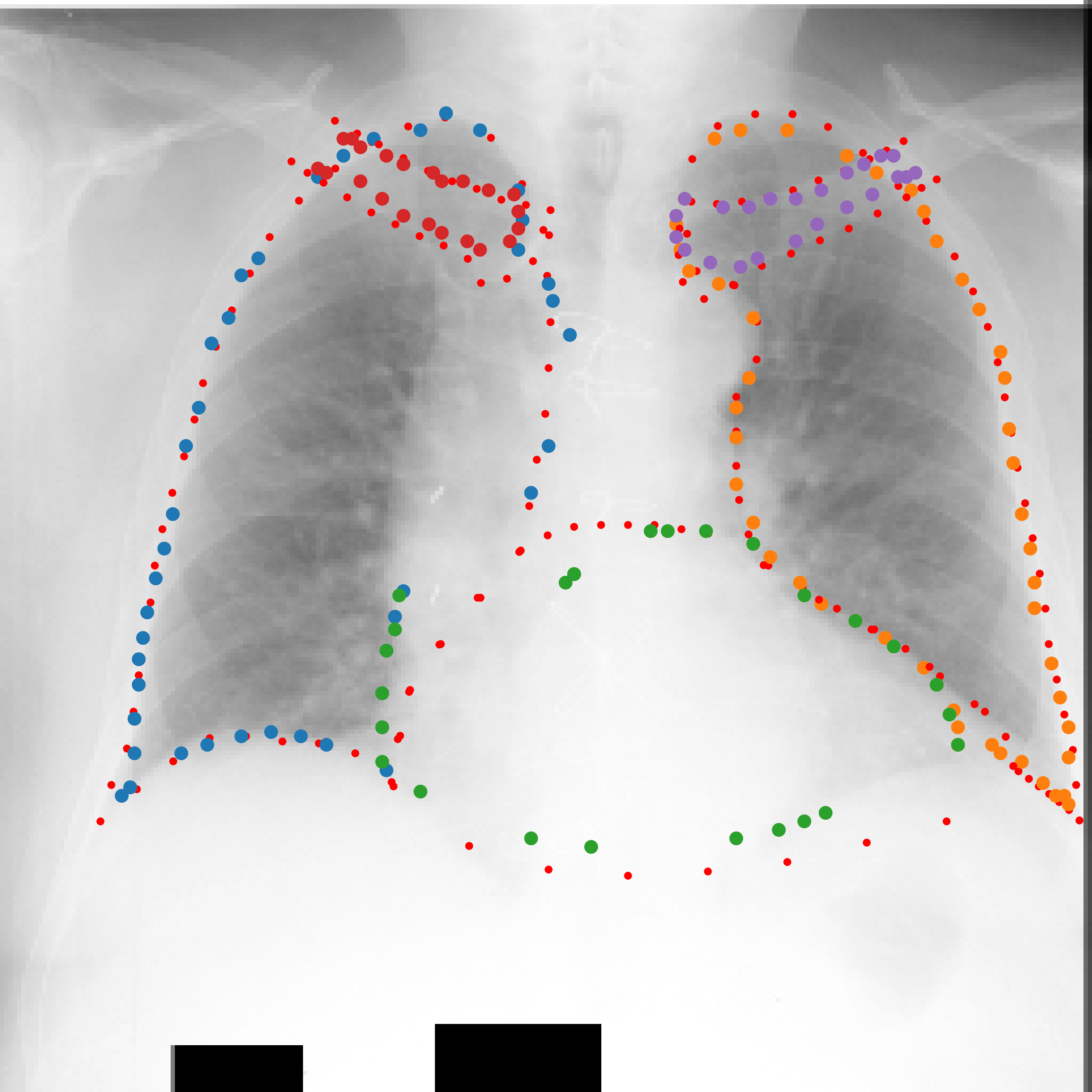}
            \caption{\begin{tabular}{rl}
                \acs{tre}: & 9.0 ± 4.2\\
                \acs{asd}: & 4.2 ± 2.3\\
            \end{tabular}}
        \end{subfigure}
    \end{subfigure}
    \begin{subfigure}{\textwidth}
        \centering
        \caption*{\textbf{ShapeFormer}}
        \begin{subfigure}{.3\textwidth}
            \includegraphics[width=\textwidth]{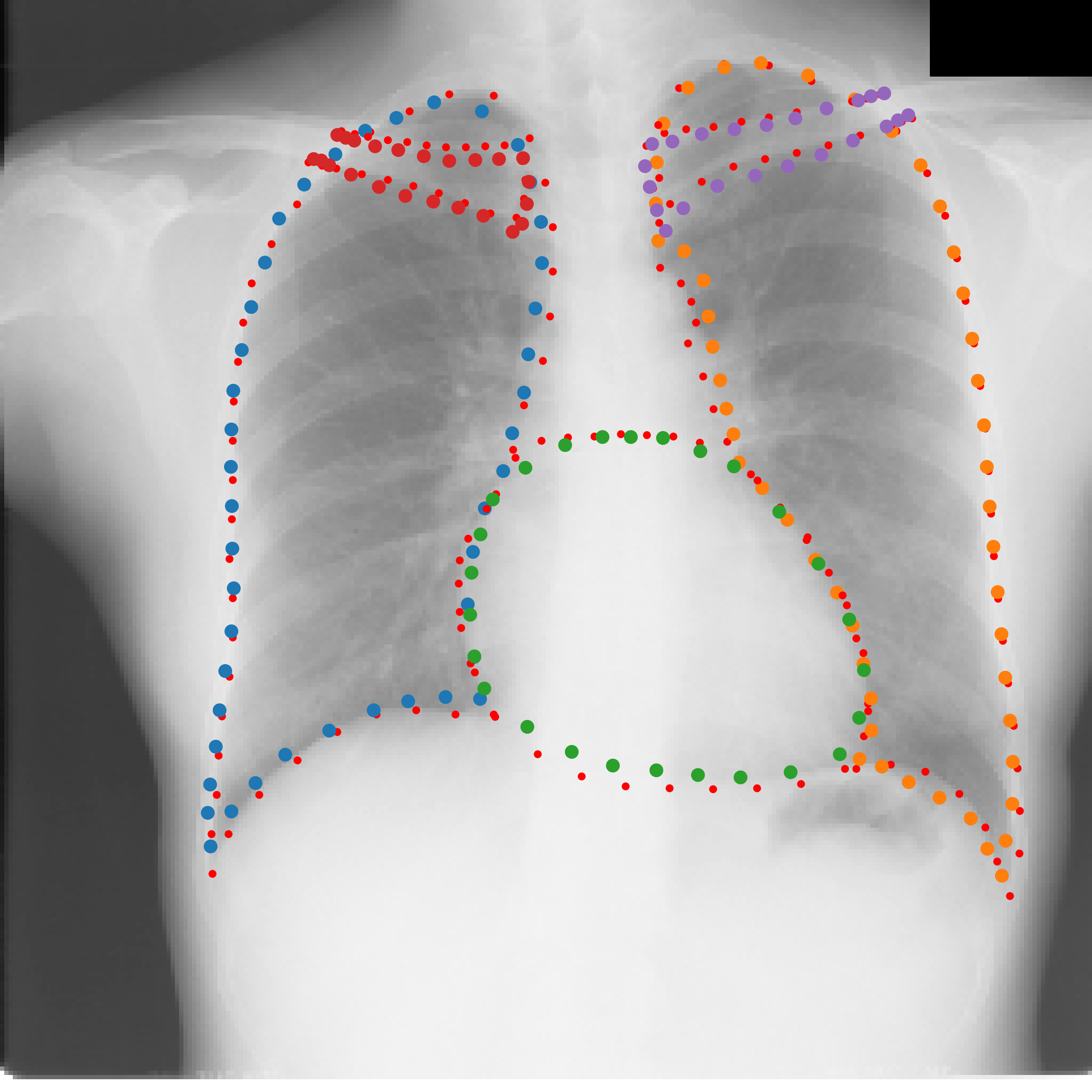}
            \caption{\begin{tabular}{rl}
                \acs{tre}: & 5.1 ± 1.3\\
                \acs{asd}: & 2.5 ± 0.6
            \end{tabular}}
        \end{subfigure}\hfill
        \begin{subfigure}{.3\textwidth}
            \includegraphics[width=\textwidth]{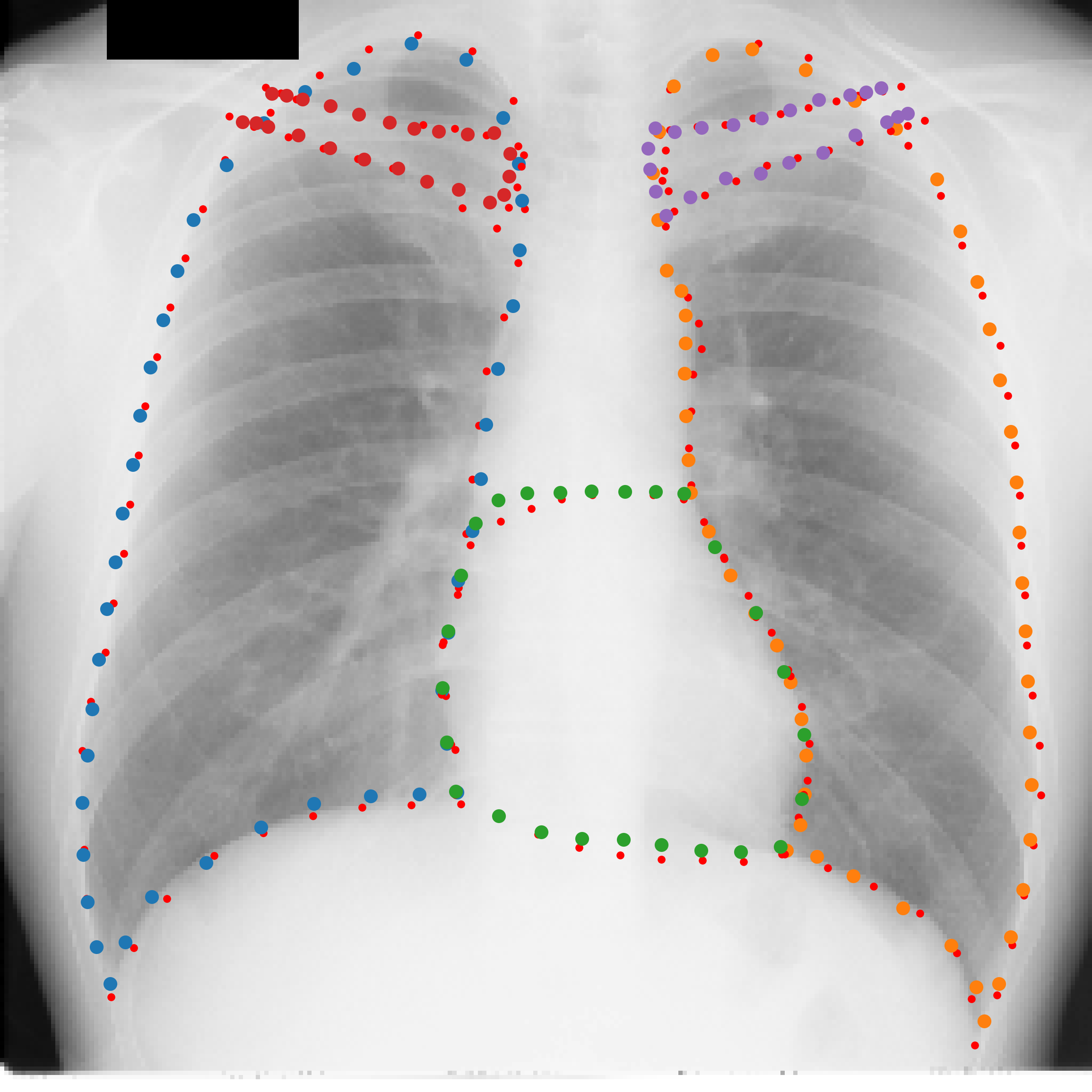}
            \caption{\begin{tabular}{rl}
                \acs{tre}: & 3.9 ± 0.7\\
                \acs{asd}: & 1.7 ± 0.3\\
            \end{tabular}}
        \end{subfigure}\hfill
        \begin{subfigure}{.3\textwidth}
            \includegraphics[width=\textwidth]{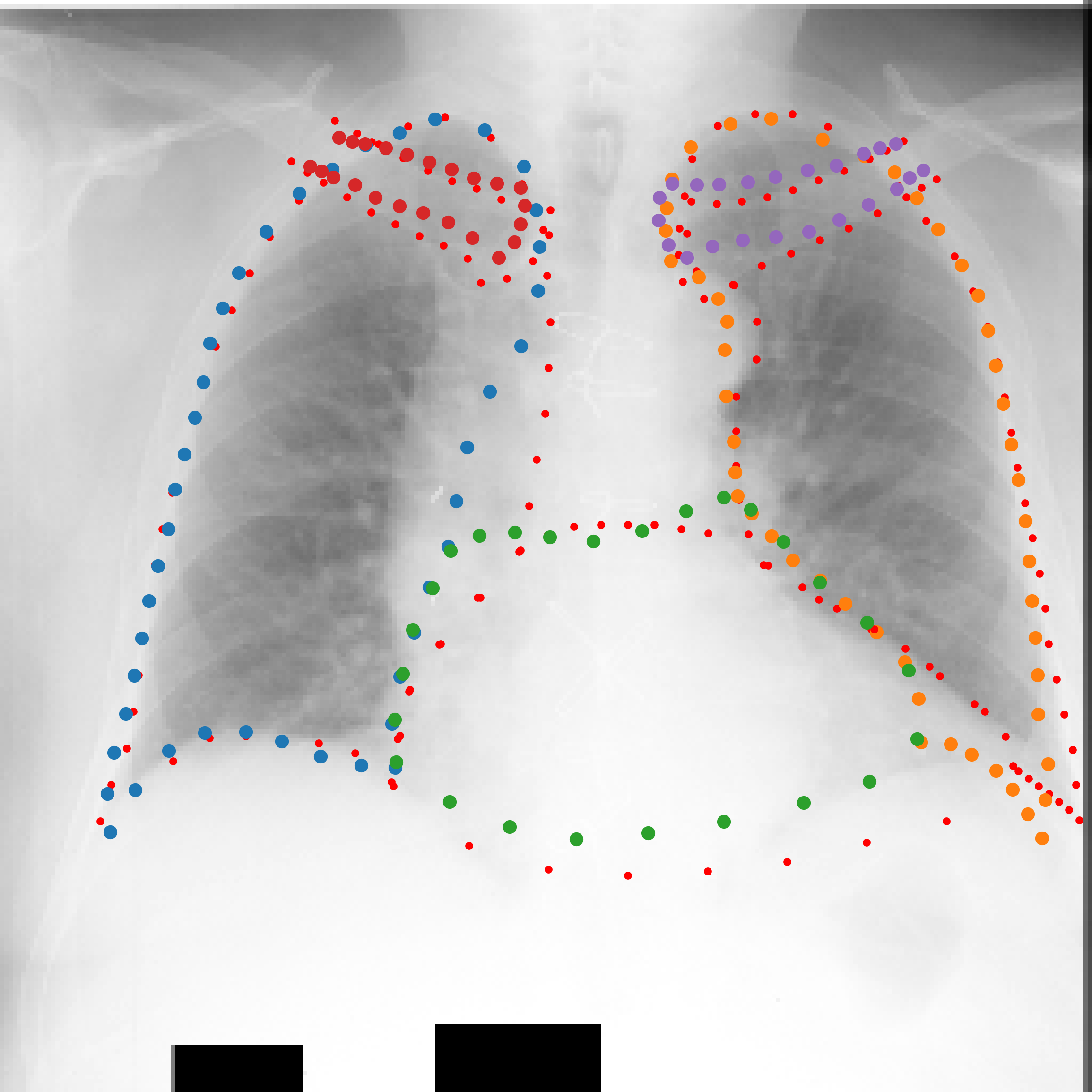}
            \caption{\begin{tabular}{rl}
                \acs{tre}: & 10.4 ± 5.1\\
                \acs{asd}: & 5.3 ± 2.1\\
            \end{tabular}}
        \end{subfigure}
    \end{subfigure}
    \caption{Qualitative result on \ac{jsrt} with the best, median and worst test case (from left to right). \acf{asd} and \acf{tre} are provided in mm. Various colors are used to distinguish different anatomical structures, with small red dots indicating the ground truth.}
    \label{fig:result_jsrt}
\end{figure}

\begin{figure}
    \centering
    \begin{subfigure}{\textwidth}
        \centering
        \caption*{\textbf{Ours}}
        \begin{subfigure}[t]{.3\textwidth}
            \includegraphics[width=\textwidth]{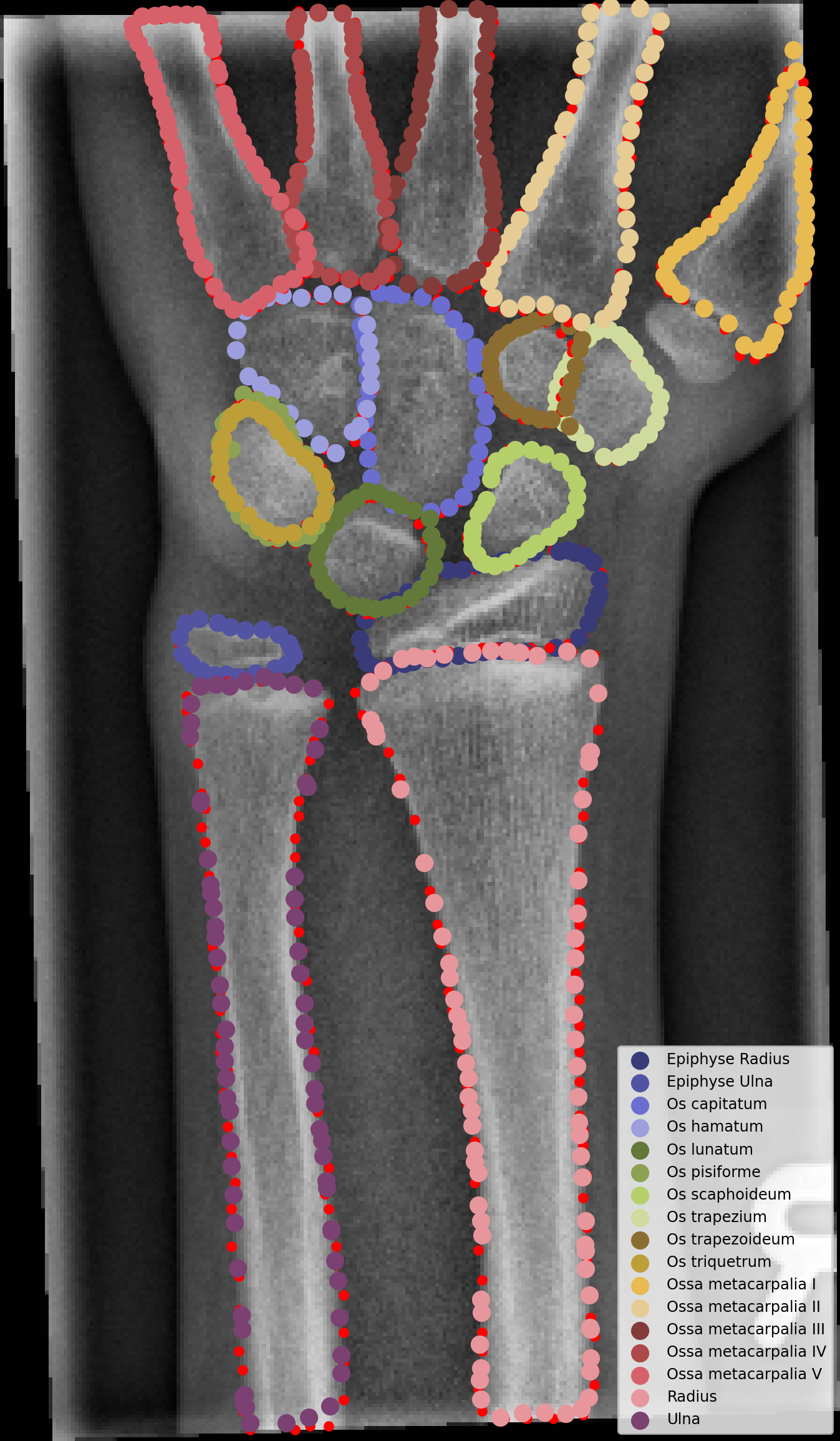}
            \caption{\begin{tabular}{rl}
                \acs{tre}: & 1.8 ± 1.2\\
                \acs{asd}: & 0.8 ± 0.2\\
            \end{tabular}}
        \end{subfigure}\hfill
        \begin{subfigure}[t]{.3\textwidth}
            \includegraphics[width=\textwidth]{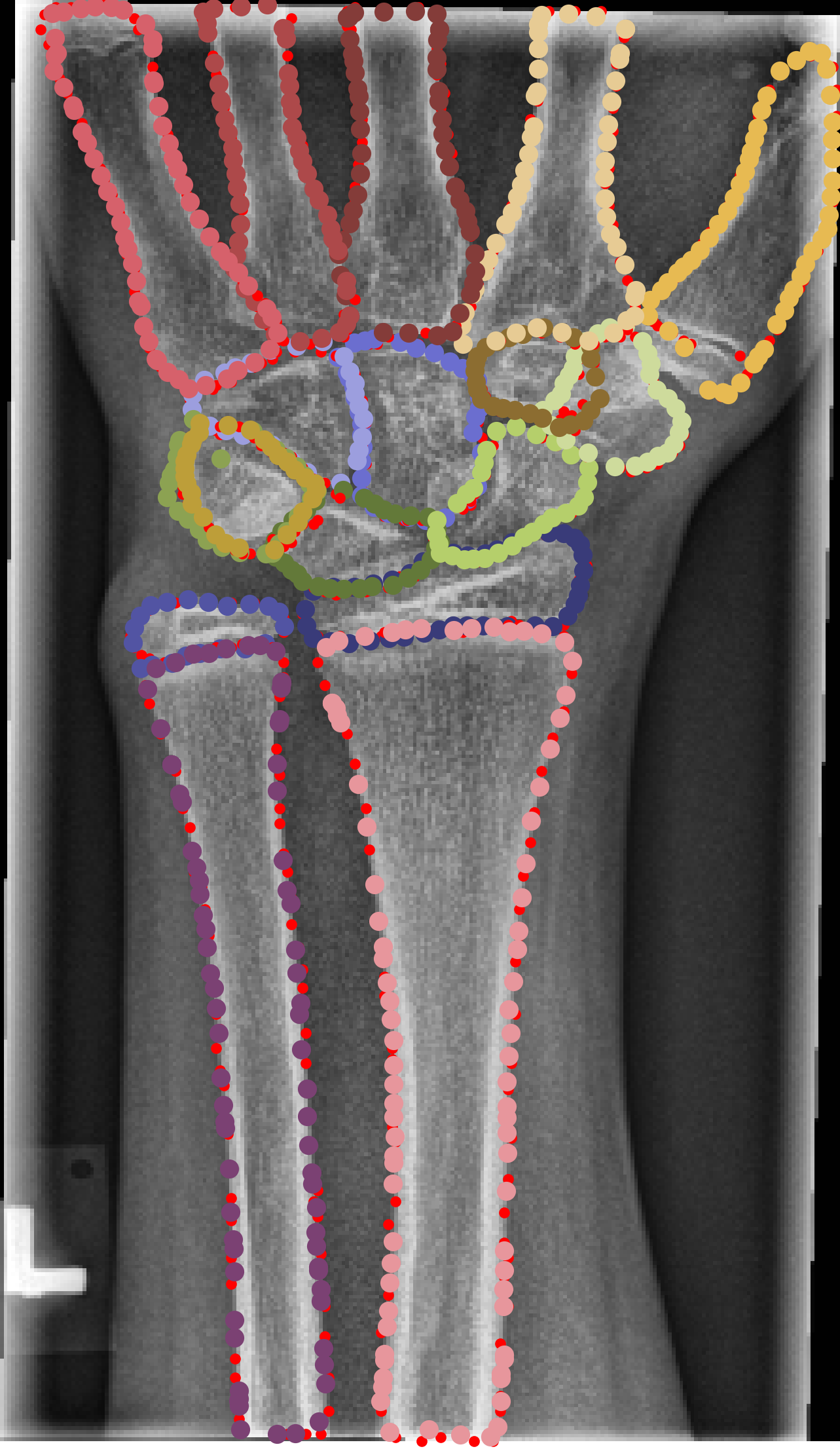}
            \caption{\begin{tabular}{rl}
                \acs{tre}: & 2.3 ± 1.2\\
                \acs{asd}: & 1.0 ± 0.3\\
            \end{tabular}}
        \end{subfigure}\hfill
        \begin{subfigure}[t]{.3\textwidth}
            \includegraphics[width=\textwidth]{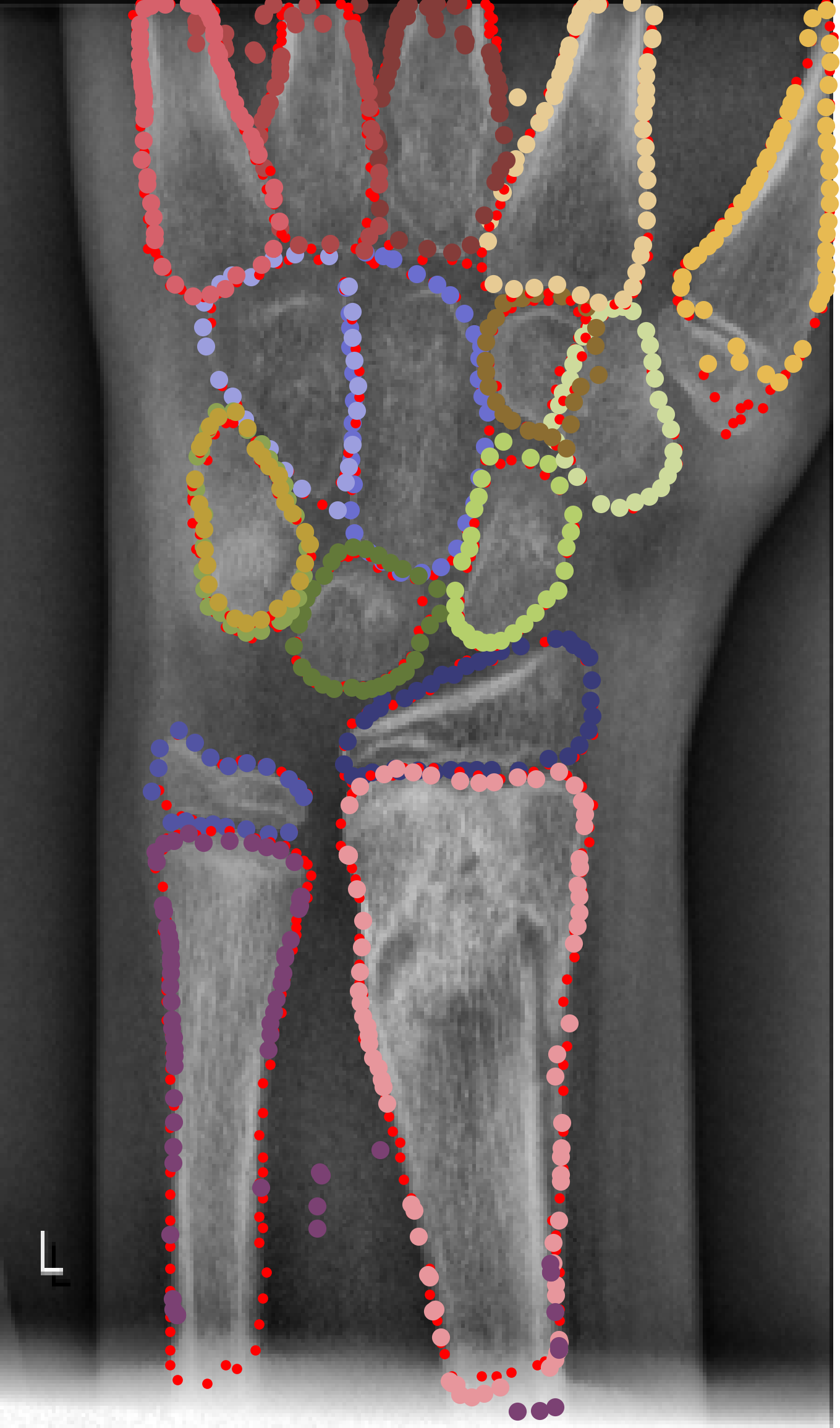}
            \caption{\begin{tabular}{rl}
                \acs{tre}: & 4.4 ± 4.2\\
                \acs{asd}: & 2.2 ± 2.9\\
            \end{tabular}}
            \label{fig:graz_uv_worst}
        \end{subfigure}
    \end{subfigure}
    \begin{subfigure}{\textwidth}
        \centering
        \caption*{\textbf{Heatmap Regression}}
        \begin{subfigure}{.3\textwidth}
            \includegraphics[width=\textwidth]{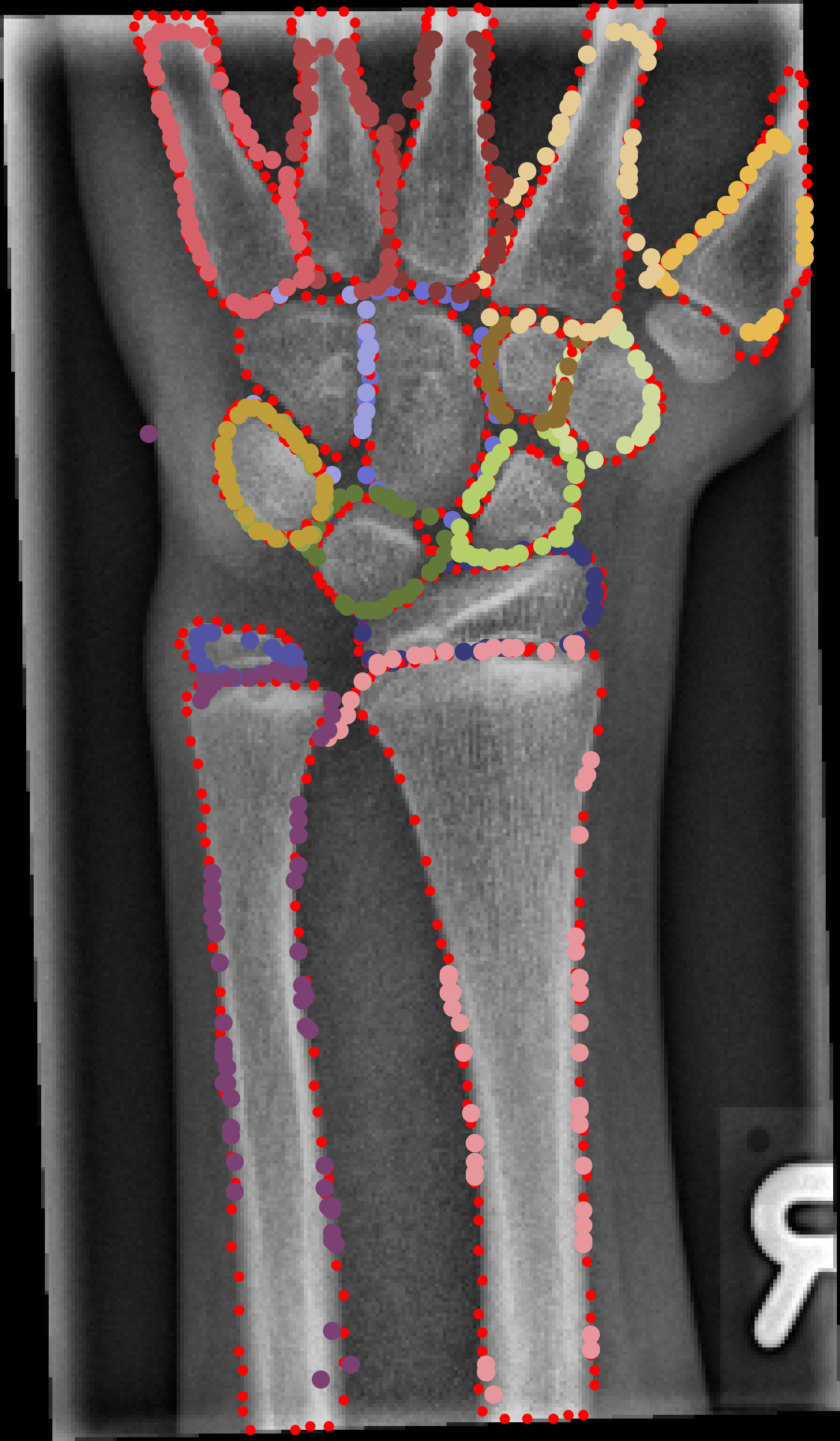}
            \caption{\begin{tabular}{rl}
                \acs{tre}: & 4.1 ± 3.8\\
                \acs{asd}: & 2.0 ± 0.4\\
            \end{tabular}}
        \end{subfigure}\hfill
        \begin{subfigure}{.3\textwidth}
            \includegraphics[width=\textwidth]{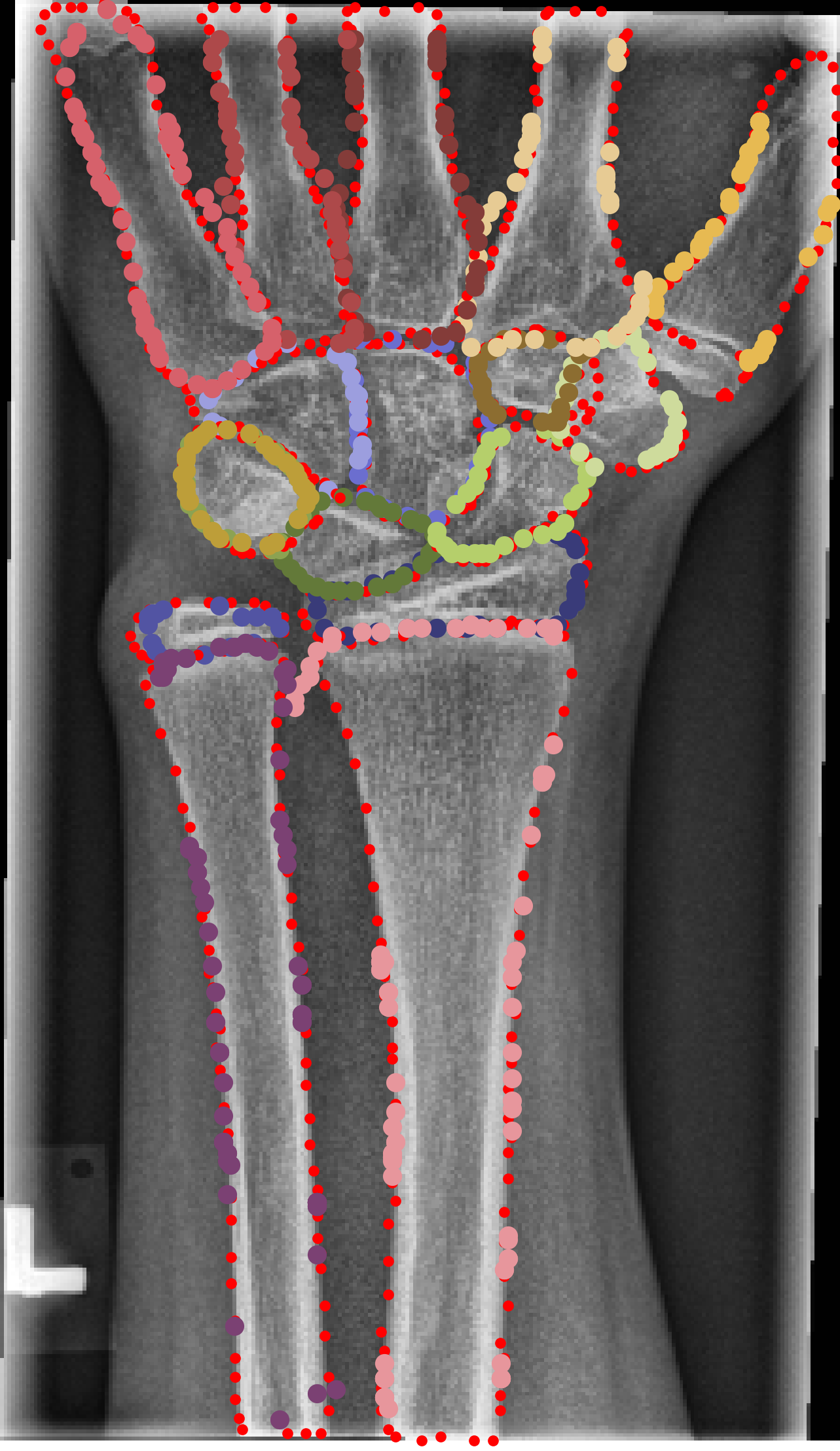}
            \caption{\begin{tabular}{rl}
                \acs{tre}: & 4.2 ± 3.1\\
                \acs{asd}: & 1.7 ± 0.3\\
            \end{tabular}}
        \end{subfigure}\hfill
        \begin{subfigure}{.3\textwidth}
            \includegraphics[width=\textwidth]{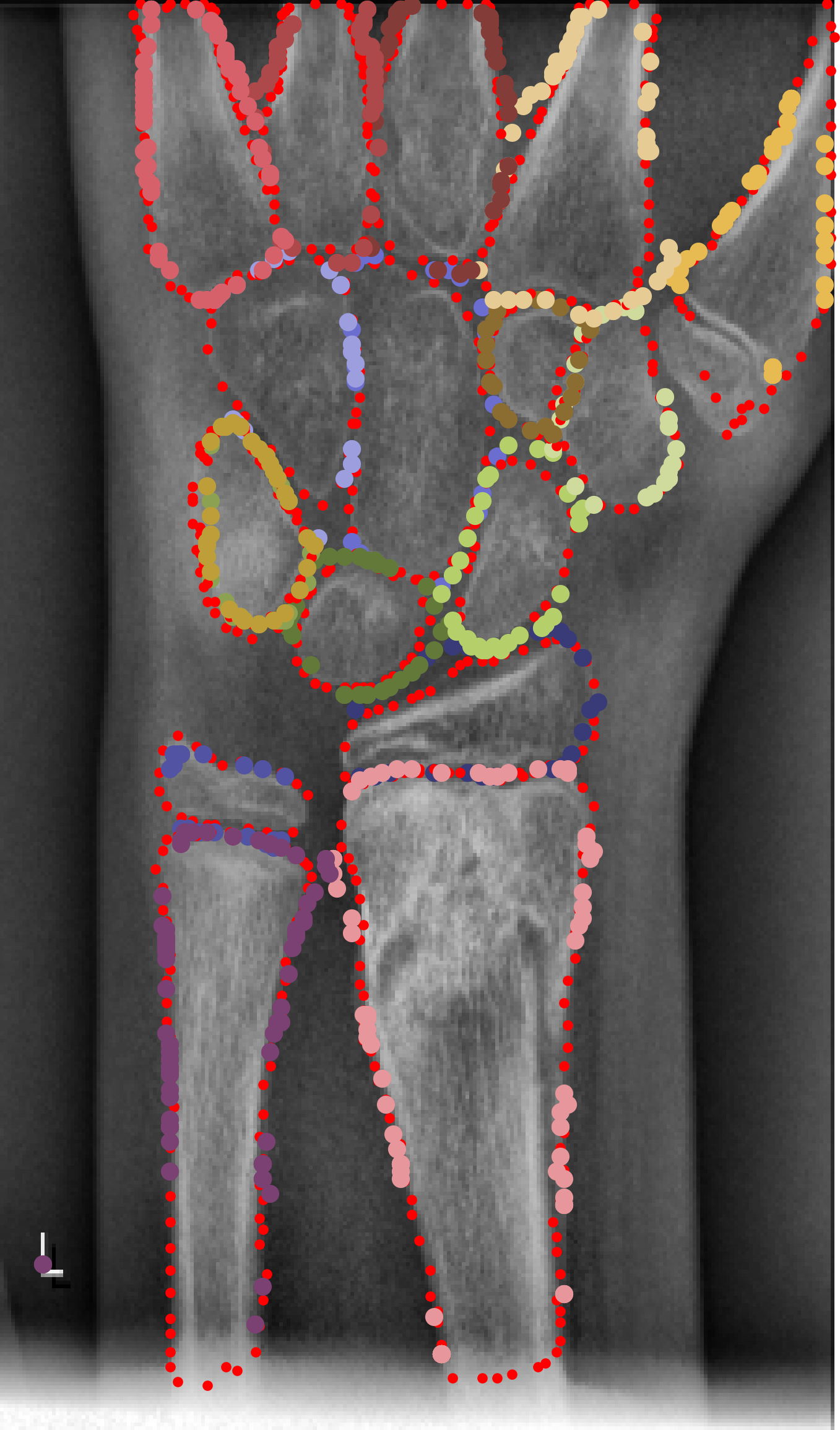}
            \caption{\begin{tabular}{rl}
                \acs{tre}: & 5.2 ± 3.1\\
                \acs{asd}: & 2.2 ± 0.5\\
            \end{tabular}}
        \end{subfigure}
    \end{subfigure}
    \caption{Qualitative result on \ac{graz} with the best, median and worst test case (from left to right). \acf{asd} and \acf{tre} are provided in pixel. Various colors are used to distinguish different anatomical structures, with small red dots indicating the ground truth.}
    \label{fig:result_graz}
\end{figure}

\begin{table}[]
    \centering
    \caption{Quantitative results on \ac{jsrt}'s test split, reported as \acs{dsc} in percentage and \acs{asd} and \acs{tre} in milimeters.}
    \begin{tabular*}{\textwidth}{l@{\extracolsep\fill}lcccccc}
        \toprule
         & Method & ShapeFormer & \acs{heatreg} & HeatRegSeg & ours & ours & nnUNet \\
        Anatomy & Metric & \cite{keuth_combining_2024} & \cite{leibe_heatmap_regression_2016} & &  & (sparse) &  \cite{isensee_nnu-net_2021} \\
        \midrule
        \multirow[t]{3}{*}{lungs} & DSC & $95.6_{\pm2.0}$ & $97.1_{\pm0.9}$ & $96.8_{\pm1.2}$ & $96.7_{\pm1.4}$ & $96.6_{\pm1.5}$ & $98.1_{\pm0.8}$ \\
         & ASD & $2.7_{\pm1.2}$ & $1.9_{\pm0.6}$ & $1.8_{\pm0.6}$ & $2.1_{\pm0.6}$ & $3.4_{\pm1.1}$ & $1.4_{\pm0.6}$ \\
         & TRE & $6.0_{\pm2.4}$ & $5.7_{\pm1.5}$ & $5.6_{\pm1.7}$ & $6.2_{\pm1.8}$ & $6.9_{\pm2.8}$ & - \\
        \midrule
        \multirow[t]{3}{*}{heart} & DSC & $92.3_{\pm4.7}$ & $94.2_{\pm2.1}$ & $92.9_{\pm3.5}$ & $92.7_{\pm2.8}$ & $92.6_{\pm2.7}$ & $95.2_{\pm2.1}$ \\
         & ASD & $4.7_{\pm2.3}$ & $3.8_{\pm1.3}$ & $3.8_{\pm1.3}$ & $3.9_{\pm1.5}$ & $5.4_{\pm1.8}$ & $3.0_{\pm1.2}$ \\
         & TRE & $9.4_{\pm9.3}$ & $8.5_{\pm8.9}$ & $8.4_{\pm9.0}$ & $8.1_{\pm9.0}$ & $9.6_{\pm8.9}$ & - \\
        \midrule
        \multirow[t]{3}{*}{clavicles} & DSC & $78.9_{\pm9.5}$ & $82.8_{\pm5.3}$ & $78.8_{\pm5.5}$ & $82.9_{\pm4.9}$ & $81.7_{\pm5.4}$ & $93.4_{\pm2.7}$ \\
         & ASD & $3.0_{\pm1.3}$ & $2.4_{\pm0.7}$ & $2.2_{\pm0.7}$ & $1.7_{\pm0.6}$ & $1.9_{\pm0.5}$ & $1.1_{\pm0.3}$ \\
         & TRE & $5.1_{\pm2.1}$ & $4.2_{\pm1.4}$ & $4.0_{\pm1.4}$ & $3.2_{\pm1.3}$ & $3.2_{\pm1.2}$ & - \\
        \midrule
        \multirow[t]{3}{*}{average} & DSC & $88.9_{\pm5.4}$ & $91.4_{\pm2.8}$ & $89.5_{\pm3.4}$ & $90.8_{\pm3.0}$ & $90.3_{\pm3.2}$ & $\mathbf{95.6_{\pm2.9}}$ \\
         & ASD & $3.5_{\pm1.6}$ & $2.7_{\pm0.9}$ & $2.6_{\pm0.9}$ & $2.6_{\pm0.9}$ & $3.5_{\pm1.1}$ & $\mathbf{1.6_{\pm1.0}}$ \\
         & TRE & $6.8_{\pm4.6}$ & $6.1_{\pm3.9}$ & $6.0_{\pm4.0}$ & $\mathbf{5.8_{\pm4.0}}$ & $6.6_{\pm4.3}$ & - \\
        \bottomrule
    \end{tabular*}
    \label{tab:result_jsrt}
\end{table}

\begin{table}[]
    \centering
    \caption{Quantitative results on \ac{graz}'s test split, reported as \acs{dsc} in percentage and \acs{asd} and \acs{tre} in pixels. The number of landmarks (lms) employed in the method is provided below the respective method.}
    \begin{tabular}{llcccc}
        \toprule
         & Method & \acs{heatreg} & \acs{heatreg} & HeatRegSeg & ours \\
        Anatomy & Metric & 720 lms  & 178 lms  & 720 lms & 720lms \\
        \midrule
        \multirow[t]{3}{*}{Carpals} & DSC & $90.1_{\pm5.0}$ & $87.0_{\pm5.1}$ & $88.0_{\pm7.8}$ & $88.8_{\pm6.9}$ \\
         & ASD & $1.8_{\pm0.4}$ & $1.8_{\pm0.5}$ & $1.8_{\pm0.4}$ & $1.0_{\pm0.4}$ \\
         & TRE & $2.5_{\pm0.7}$ & $2.7_{\pm0.8}$ & $2.5_{\pm0.7}$ & $1.7_{\pm0.6}$ \\
        \midrule
        \multirow[t]{3}{*}{Metacarpals} & DSC & $87.1_{\pm5.3}$ & $83.8_{\pm5.2}$ & $94.7_{\pm3.0}$ & $94.4_{\pm3.4}$ \\
         & ASD & $2.0_{\pm0.4}$ & $2.0_{\pm0.7}$ & $2.0_{\pm0.4}$ & $1.2_{\pm0.9}$ \\
         & TRE & $4.8_{\pm1.7}$ & $5.6_{\pm1.9}$ & $5.0_{\pm1.9}$ & $2.5_{\pm1.5}$ \\
        \midrule
        \multirow[t]{3}{*}{Ulna/Radius} & DSC & $87.2_{\pm8.8}$ & $81.8_{\pm9.7}$ & $94.3_{\pm4.8}$ & $94.4_{\pm4.6}$ \\
         & ASD & $1.9_{\pm0.6}$ & $1.6_{\pm0.4}$ & $1.7_{\pm0.6}$ & $1.0_{\pm0.5}$ \\
         & TRE & $6.9_{\pm4.6}$ & $5.7_{\pm3.2}$ & $6.9_{\pm4.3}$ & $3.2_{\pm2.2}$ \\
        \midrule
        \multirow[t]{3}{*}{average} & DSC & $88.1_{\pm6.4}$ & $84.2_{\pm6.7}$ & $92.3_{\pm5.2}$ & $\mathbf{92.6_{\pm5.0}}$ \\
         & ASD & $1.9_{\pm0.5}$ & $1.8_{\pm0.5}$ & $1.8_{\pm0.5}$ & $\mathbf{1.1_{\pm0.6}}$ \\
         & TRE & $4.8_{\pm2.3}$ & $4.6_{\pm2.0}$ & $4.8_{\pm2.3}$ & $\mathbf{2.5_{\pm1.4}}$ \\
        \bottomrule
    \end{tabular}
    \label{tab:results_graz}
\end{table}

Due to the lack of native segmentation estimation for the ShapeFormer and \ac{heatreg}, we apply the necessary conversions as post-processing step to enable the comparison with the \acf{dsc}.
Please note, this could lead to some loss of performance while transferring the (subpixel) landmark position to a mask.
Likewise for the nnUNet, providing only segmentation masks, we calculate the \acf{asd} based on all contour points instead of the landmarks like in the other methods.

\subsection{Results on \ac{jsrt}}
Tab. \ref{tab:result_jsrt} shows with its quantitative results on the \ac{jsrt}'s test split all methods outperforming our previous, shape-based work.
In context of \ac{dsc} and \ac{asd} the nnUNet yields the overall best result (\ac{dsc} of $95.6\pm2.9\,\%$ and \ac{asd} of $1.6\pm1\,\text{mm}$), which introduces a gap of over $4\,\%$ and $1\,\text{mm}$ to our method and \ac{heatreg}.
However, in context of the \ac{tre}, the nnUNet lacks any landmark detection and our method provides the overall best performance with an error of $5.8\pm4\,\text{mm}$ vs $6.1\pm3.9\,\text{mm}$ for \ac{heatreg}.\par
Considering the scores of the individually anatomical structures, the bias of the \ac{dsc} favouring large structures becomes visible, supporting the use of \ac{asd}.
In terms of \ac{tre} and \ac{asd} representing the performance of the landmark detection, the contrast of the anatomical structures matters, resulting in higher landmark accuracy for structures with a higher contrast.
This puts the clavicles as bones first, followed by the lungs, where the pleural cavity already smooth their boundaries, while the heart producing the highest error due to its low contrast boundaries, making it difficult to detect the exact position of its landmarks.\par
As ablation studies, we first investigate the importance of dense supervision provided with the ground truth $uv$-map by omitting $\mathcal{L}_\varphi$ from Eq. \ref{eq:loss_function}.
With this, the model has to interpolate the $uv$ values between the sampling points provided on the ground truth landmarks' position on its own.
However, when we use only this kind of sparse supervision, the performance drops to the level of the ShapeFormer.
Secondly, the extension of the \ac{heatreg} approach with a segmentation head (HeatRegSeg) shows no benefit in the \ac{jsrt} setting, where the precise landmark prediction of \ac{heatreg} ($6.1\,\text{mm}$ vs HeatRegSeg $6\,\text{mm}$) already allows the extraction of accurate segmentations (\ac{dsc} $91.4\,\%$ vs $89.5\,\%$).\par
Fig. \ref{fig:result_jsrt} compares qualitatively results of our method, \ac{heatreg} and ShapeFormer.
We chose the best (left column), the median (middle column) and worse (right column) test case regarding the \ac{asd}.
While our method shows accurate landmark detection for the best case, especially for the clavicles and lung tips, the median case point out a potential drawback.
While our method has the advantage of concurrently generating a segmentation mask, this integrated learning with the $uv$ mapping can also lead to complications. For instance, if the heart's segmentation is overestimated (Fig. \ref{fig:jsrt_heart_overestimation}), it can consequently cause an overestimation in the $uv$-map. This is due to the learned anatomical correspondences between the $uv$ coordinates and the segmentation mask, which can affect each other.
For challenging cases with low contrast, the implicitly and explicitly learned geometric shape by the \ac{heatreg} and ShapeFormer results in a more plausible prediction and thus lower error.
This holds especially true for the worst example, where the ability of shape-based methods generating only anatomically plausible prediction can show its strength, as it has already been demonstrated in our previous work \cite{keuth_combining_2024}.


\subsubsection{Addition of New Landmarks without Retraining}
For evaluating the performance of four newly added landmarks without retraining ($\textcolor{orange}{\blacklozenge}$ markers in $\mathbf{T}$ in Fig. \ref{fig:uv_generation}), we compare the \ac{tre} results of our model on the test split against the mean shape (values shown in parentheses).
For the four clavicle-related points, our model achieves a \ac{tre} of $20.5\pm5.1$ ($15\pm7.1$) mm.
The \ac{tre} for the center point is $8.3\pm4.2$ ($17.5\pm9.1$) mm, for the right contour point $7.3\pm5.9$ ($19.5\pm10$) mm, and for the center of the lung apex $7\pm5.4$ ($29.3\pm15.4$) mm.
These results demonstrate that the $uv$-based paradigm enables the prediction of landmarks not seen during training, achieving comparable accuracy to the known landmarks ($5.8\pm\,\text{mm}4$ "ours" in Tab.\ref{tab:result_jsrt}).
However, the results highlight a limitation for the clavicle landmarks, where our model performs worse than the mean shape.
This suggests that the newly added landmarks must correspond anatomically to the object's (in this case, the left lung's) $uv$ map with their transformation to remain relative to the object's geometry.
This is not satisfied for the clavicles in the \ac{jsrt} dataset, where the X-rays capture chests in varying breathing states, causing the clavicles to "move" relative to the upper lungs, disrupting the necessary anatomical alignment.

\subsection{Results on \ac{graz}}
In the more complex setting of the \ac{graz} dataset with 720 landmarks of 17 different bones, including the small carpals and epiphyses, our method outperforms \ac{heatreg} by almost halving the \ac{tre} ($2.5\pm1.4$ vs $4.8\pm2.3$) and \ac{asd} ($1.1\pm0.6$ vs $1.9\pm0.5$) (see. Tab. \ref{tab:results_graz}).
When there is a deficiency in accurate landmarks, the segmentation maps derived from these landmarks also tend to be imprecise, resulting in a lower \ac{dsc} of $88.1\pm6.4\,\%$ compared to our method ($92.6\pm5\,\%$).
However, predicting the segmentations directly via the dedicated segmentation head (HeatRegSeg) yields a \ac{dsc} of $92.3\,\%$ regardless of its imprecise landmarks.\par
Since we leave the underlying UNet architecture unchanged, projecting 720 heatmaps from a latent space with 64 dimension could introduce a bottleneck.
To test if the model capacity is limiting the performance, we employ farthest point sampling and reduce the 720 initial landmarks to a comparable range of the 166 \ac{jsrt}'s landmarks reconstructing a setting, where \ac{heatreg} has reached good results.
In order to preserve the original distribution of landmarks for each bone, we do not establish a fixed total number of landmarks. Instead, we use 25\% of the landmark count associated with each individual bone, ensuring that the landmark distribution remains proportional to the original distribution. This results in a cumulative total of 178 landmarks.
However, as shown in Tab. \ref{fig:result_graz} (\acs{heatreg} 178 lms), this only slightly improves performance (\ac{tre}: $4.6\pm2$ vs. $4.8\pm2.3$, and \ac{asd}: $1.8\pm0.5$ vs. $1.9\pm0.5$) concluding that our method is superior in this setting.
Moreover, it highlights the advantage of the $uv$ paradigm, which allows for predicting an arbitrary number of landmarks for each anatomy without needing to modify the U-Net architecture, where \ac{heatreg} requires the introduction of a channel and thus additional parameters for each landmark.\par
Fig. \ref{fig:result_graz} shows that our method predicts accurate landmarks for the carpal bones and epiphysis of radius and ulna.
While for the most cases this also holds true for the metacarpals, radius and ulna, our method can produce inaccurate landmarks for these bones, if the image shows a rather smaller section of the wrist (see Fig. \ref{fig:graz_uv_worst}).
However, the landmarks of carpal bones and epiphysis remain highly accurate.
Considered the qualitative result of the \ac{heatreg} in the second row, it produces high errors in all three images leaving areas/borders of many bones, especially radius, ulna and ossis hamatum uncovered.

\subsubsection{Bone Age Regression with Landmarks}
\begin{figure}
    \centering
    \begin{subfigure}[c]{.4\textwidth}
        \includegraphics[width=\textwidth, trim={0 1.1cm 4cm 0cm}, clip]{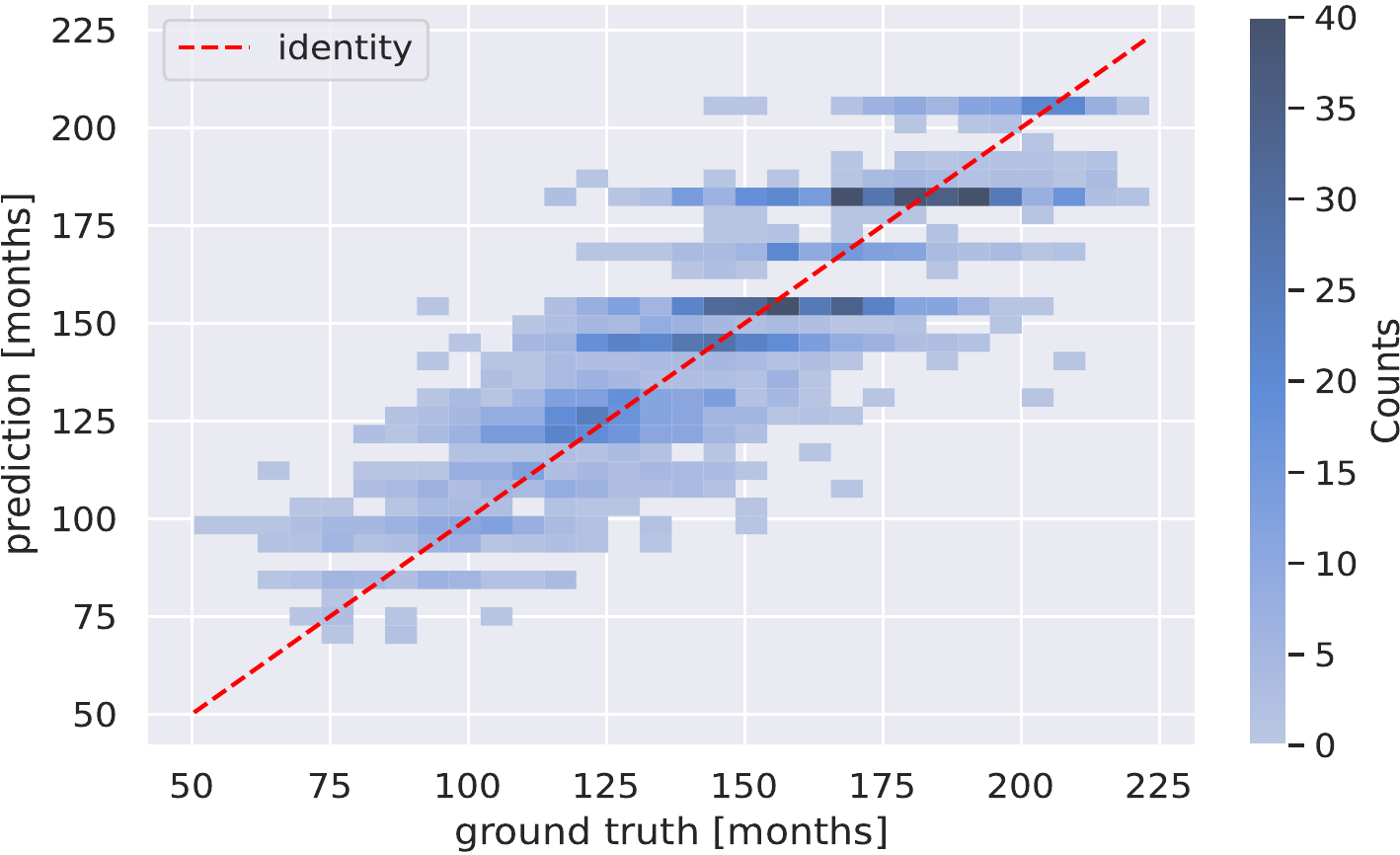}
    \end{subfigure}\hspace{.5cm}
    \begin{subfigure}[c]{.5\textwidth}
        \begin{tabular}{lcc}
            \toprule
            Landmarks by & MAE [month] & R² \\
            \midrule
            ground truth & 14.1 & 0.708\\
            ours & \textbf{13.9} & \textbf{0.715}\\
            \acs{heatreg} \cite{leibe_heatmap_regression_2016} & 14.1 & 0.713\\
            \midrule
            mean age (train) & 27.4 & 0\\
            \bottomrule
        \end{tabular}
    \end{subfigure}
    \caption{Results for bone age regression with landmarks provided by different methods. Left: the error plot when using landmarks generated by our method. Right: mean absolute age estimation error (MAE) and $R^2$ correlation score.}
    \label{fig:bone_age_regression}
\end{figure}

Fig. \ref{fig:bone_age_regression} shows the result of the bone age regression task as an example for a clinical application of landmark detection.
We achieved results of a mean absolute age estimation error (MAE) of 13.9 months with an $R^2$ correlation score of 0.715 for the proposed method that are comparable to the results obtained for the ground truth landmarks MAE$=$14.1 months ($R^2=0.708$) and the heatmap-based estimation with MAE$=$14.1 months ($R^2=0.713$).
While there is a certain amount of ambiguity between chronological age (used as ground truth in this case) and biological age, our results demonstrate that landmark-based bone age regression provides a viable automatic solution, but is also less sensitive to spurious landmark errors.

\section{Conclusion}
In our work, we proposed a dense image-to-shape representation that enables landmark detection and can be trained along with semantic segmentation utilizing state-of-the-art \ac{cnn} architecture for segmentation. 
While our method is on pair with learning-based state-of-the-art landmark detection approaches in a X-ray thorax setting, it outperforms them in the more challenging setting of paediatric wrist, including 17 bones and 720 landmarks.
Due to our formulation of landmark detection via dense representation, our method does not explicitly train on the landmarks themselves, allowing the inclusion of new landmarks without the need of retraining.
Our work demonstrates the applicability of the $uv$ paradigm for radiographs, and while similar approaches have been successfully deployed in natural image processing \cite{Gler2016DenseRegFC, Gler2018DensePoseDH}, the applicability to further medical domains yet remains to be explored.




\section*{Declarations}
The authors have no relevant financial or non-financial interests to disclose.

\subsection{Code and Data availability}
Our code, along with our generated annotations (landmarks and bone segmentation) for the \ac{graz} dataset, is openly accessible to the public at \url{https://github.com/MDL-UzL/DenseSeg}. 

\subsection{Funding}
This research has been funded by the state of Schleswig-Holstein, Grant Number 220 23 005.





\bibliography{sn-bibliography}

\end{document}